\documentclass{article}

\PassOptionsToPackage{numbers}{natbib}
\usepackage[final]{neurips_2025}
\usepackage{graphicx}
\usepackage{subcaption}
\usepackage{makecell}
\usepackage{wrapfig}

\usepackage{colortbl}
\usepackage{xcolor}

\usepackage{algorithm}
\usepackage{algpseudocode}

\usepackage[utf8]{inputenc} % allow utf-8 input
\usepackage[T1]{fontenc}    % use 8-bit T1 fonts
\usepackage{hyperref}       % hyperlinks
\usepackage{url}            % simple URL typesetting
\usepackage{booktabs}       % professional-quality tables
\usepackage{amsfonts}       % blackboard math symbols
\usepackage{nicefrac}       % compact symbols for 1/2, etc.
\usepackage{microtype}      % microtypography
\usepackage{xcolor}         % colors
\usepackage{multirow}
\usepackage[most]{tcolorbox}

\title{MUSTAFAR: 
Pro\underline{m}oting \underline{U}n\underline{st}ructured Sp\underline{a}rsity \underline{f}or KV C\underline{a}che Pruning in LLM Infe\underline{r}ence}
\author{%
    \small \textbf{Donghyeon Joo$^{1}$, Helya Hosseini$^{1}$, Ramyad Hadidi$^{2}$, Bahar Asgari$^{1}$} \\
    \small$^{1}$Department of Computer Science, University of Maryland, $^{2}$d-Matrix \\
    \small \texttt{\{dhjoo98,helia,bahar\}@umd.edu}, \texttt{rhadidi@d-matrix.ai}
}

\begin{document}

\maketitle

\begin{abstract}

We demonstrate that unstructured sparsity significantly improves KV cache compression for LLMs, enabling sparsity levels up to 70\% without compromising accuracy or requiring fine-tuning. We conduct a systematic exploration of pruning strategies and find per-token magnitude-based pruning as highly effective for both Key and Value caches under unstructured sparsity, surpassing prior structured pruning schemes. The Key cache benefits from prominent outlier elements, while the Value cache surprisingly benefits from a simple magnitude-based pruning despite its uniform distribution. 
KV cache size is the major bottleneck in decode performance due to high memory overhead for large context lengths. 
To address this, we use a bitmap-based sparse format and a custom attention kernel capable of compressing and directly computing over compressed caches pruned to arbitrary sparsity patterns, significantly accelerating memory-bound operations in decode computations and thereby compensating for the overhead of runtime pruning and compression.
%
%Our sparse format coupled with a custom attention kernel
Our custom attention kernel coupled with the bitmap-based format delivers substantial compression of KV cache up to 45\% of dense inference and thereby enables longer context lengths and increased tokens/sec throughput of up to 2.23$\times$ compared to dense inference. 
%This advancement not only facilitates longer context lengths but also democratizes access to high-performing LLMs, enabling efficient deployment. 
Our pruning mechanism and sparse attention kernel is available at \url{https://github.com/dhjoo98/mustafar}.
\end{abstract}

\section{Introduction}
%Part1: Significance of KV cache 
In the age of Large Language Models (LLMs), advances in the machine learning domain~\cite{vaswani2017attention, ainslie2023gqa, flashattention2} and the fast and efficient computing systems~\cite{jouppi2023tpuv4, modelingstc} have led to the emergence of highly capable LLMs that can summarize a book~\cite{kim2024fables}, write a compelling story~\cite{huot2025agentsroom}, code a library~\cite{zhao2024commit0}, and generally reason over longer contexts than ever before~\cite{deepseek2025r1}.   
As LLMs are increasingly tasked with processing longer sequences, the memory overhead associated with key-value (KV) caching has emerged as a critical bottleneck to scaling context length.

Prior work has approached the challenge of KV cache memory overhead through techniques such as quantization~\cite{kivi, zipcache, 1bitkv, qhitter}, low-rank approximation~\cite{asvd, palu, sun2025shadowkv, zhang2025tensorproduct, lin2024matryoshkakv}, token-wise eviction~\cite{h2o, scissorhands, snapkv, less, keyformer, ge2024discard}, and structured pruning (e.g., channel-wise removal~\cite{think, lv2024kvpruner}). 
The need to improve individual compression techniques has become increasingly important, especially as joint applications of multiple methods, such as pruning combined with token eviction~\cite{think}, quantization with token-wise eviction~\cite{qhitter}, and low-rank approximation with quantization~\cite{palu}, gain popularity.
%However, previous works on KV cache pruning have been limited to structured pruning due to the difficulty in leveraging a finer granularity of sparsity efficiently during execution. 
However, previous work on KV cache pruning have been limited to structured pruning, primarily due to the difficulty of efficiently leveraging finer-grained (i.e., unstructured) sparsity during execution.
Effective pruning of the KV cache entails two core challenges: (1) achieving substantial reduction in KV cache size while preserving model accuracy, and (2) ensuring that the runtime pruning and compression processes are sufficiently efficient (i.e., the associated overhead must not outweigh the latency gains introduced by the resulting sparsity).

\begin{comment}
\begin{figure}[htbp]
\centerline{\includegraphics[width=0.80\textwidth]
{figs/top_level.pdf}}
\caption{High-Level Overview of Mustafar. Green region describes the Pruning Algorithm of Section~\ref{sec:2}, Red region describes the custom Sparse Attention Kernel of Section~\ref{sec:3}}
\label{fig:top_fig}
\end{figure}
\end{comment}

\begin{wrapfigure}{r}{0.5\textwidth}  
    \centering
    \includegraphics[width=0.50\textwidth]{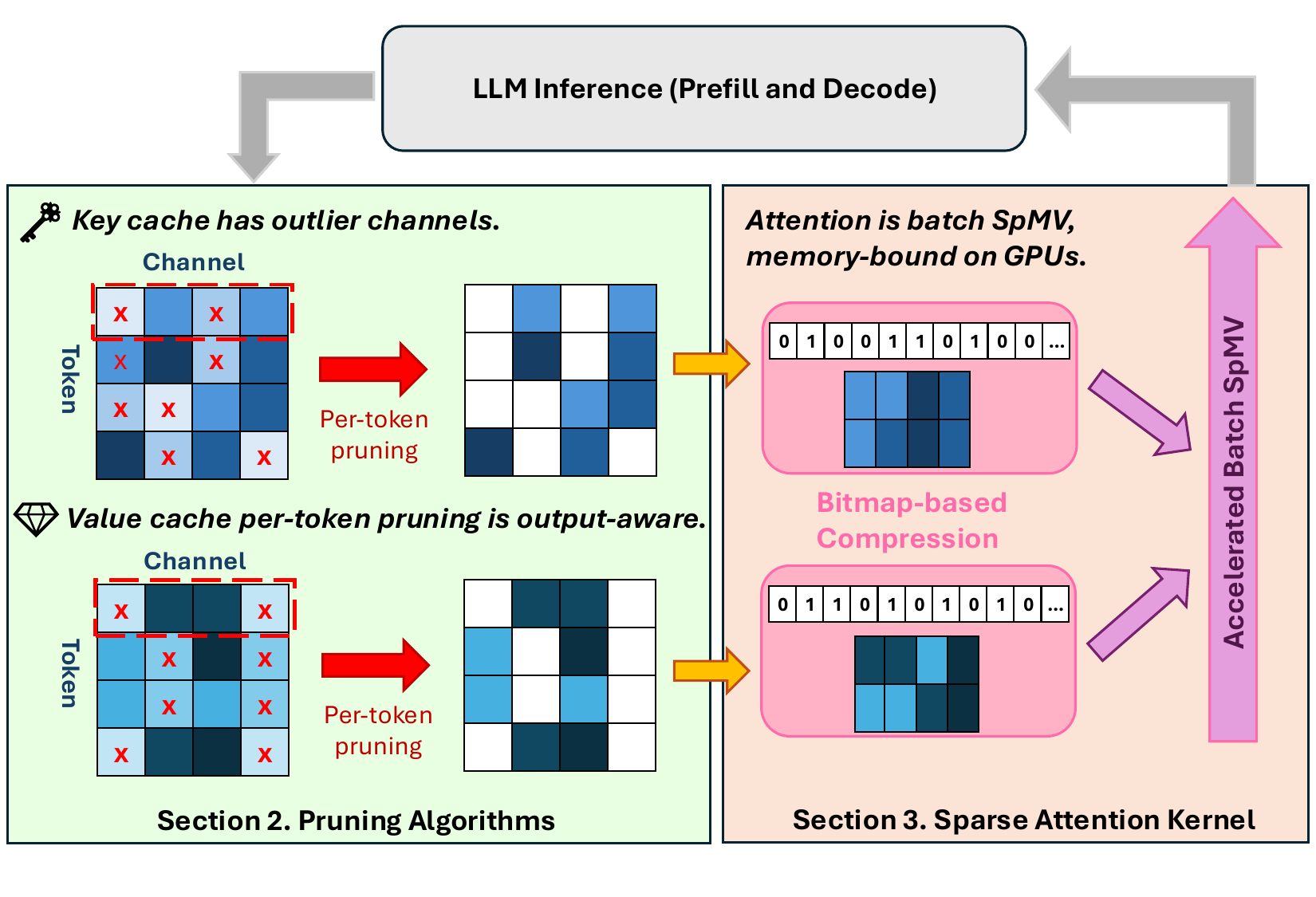}
    \caption{High-level overview of Mustafar. Green region describes the pruning algorithm of Section~\ref{sec:2}, pink region describes the custom sparse attention kernel of Section~\ref{sec:3}.}
    \label{fig:top_fig}
\end{wrapfigure}

In this paper, we find that removing any constraint on the sparsity pattern, effectively unstructured sparsity can ensure that compressed KV cache perform with minimal model accuracy degradation while being pruned to a higher sparsity. 
In Section~\ref{sec:2} (green region of Figure~\ref{fig:top_fig}), we first present our journey to find the optimal pruning algorithm for the key and value cache, based on the element magnitude distributions of the KV cache. We explore the feasibility of various pruning algorithms on both KV cache to conclude that applying a simple per-token magnitude-based pruning on both Key and Value caches is capable of preserving the model accuracy at a high sparsity, while also demonstrating strong compatibility with orthogonal compression techniques.

Section~\ref{sec:3} (pink region of Figure~\ref{fig:top_fig}) discusses the next step: having induced sparsity in the KV cache, the challenge becomes leveraging the unstructured sparsity to reduce memory footprint and accelerate computation.
To this end, we adopt a bitmap-based sparse format that serves two purposes. First, the bitmap enables maximal compression of matrices with arbitrary sparsity patterns. Second, this maximal compression of matrix operands translates into computational speedup of the attention operation, which is severely memory-bound on GPUs. Alongside the sparse format, we introduce the custom attention kernel tailored to operate on the bitmap-based sparse format. We see that the speedup of our attention kernel overshadows the latency introduced by runtime pruning and compression, meanwhile effectively compressing the KV cache to high sparsity with minimal accuracy degradation. 

In summary, we demonstrate that adopting unstructured sparsity in the KV cache without imposing constraints on the pruning pattern enables higher degrees of sparsity while preserving model accuracy. Furthermore, we introduce the necessary computational tools to support unstructured sparsity efficiently, ensuring that the derived high sparsity leads to gains in memory compression and end-to-end inference throughput. %computational efficiency.

\section{Pruning Algorithm for Unstructured Sparsity}
\label{sec:2}

\begin{tcolorbox}[colback=blue!5!white, colframe=blue!5!white]
\textbf{Question}: Does removing structural constraints in KV cache pruning allow for higher sparsity while preserving model accuracy more effectively than structured pruning methods?
\end{tcolorbox}

We explore the potential unstructured sparsity on KV cache pruning by considering the two factors for Key and Value cache pruning: pruning direction and output-awareness. \textbf{Pruning Direction} refers to the axis along which sparsity is induced when selecting elements for removal. Since both the Key and Value caches are represented as matrices with dimensions $[tokens \times channels]$, we consider two primary pruning directions: per-channel pruning, which determines target sparsity across each channel (i.e., across tokens for each channel), and per-token pruning, which determines target sparsity across each token's cache (i.e., across model dimensions for each token).
\textbf{Output-Awareness} refers to the use of a scoring metric that serves as a proxy for estimating each element’s contribution to the operation’s output. 
Commonly employed in LLM weight pruning~\cite{wanda} and structured KV cache pruning~\cite{think}, this technique involves computing a score for each pruning unit such as a channel or an element by taking the product of the corresponding element with its associated input. This approach effectively captures the element’s influence on the final output, guiding more informed pruning decisions. For a fair and effective comparison between pruning strategies, we uniformly employ a \textbf{local dense window}, where the recent 32 tokens remain untouched during the decode phase. Previous works~\cite{h2o, think} have shown that this is effective in preserving model accuracy, meanwhile being small enough in size to not severely impact the compression. 

\subsection{Pruning Key Cache}

\begin{figure}[h]
\centering
\begin{subcaptionbox}{Magnitude distribution of Key cache\label{fig:sub1}}[0.45\textwidth]
  {\includegraphics[width=\linewidth]{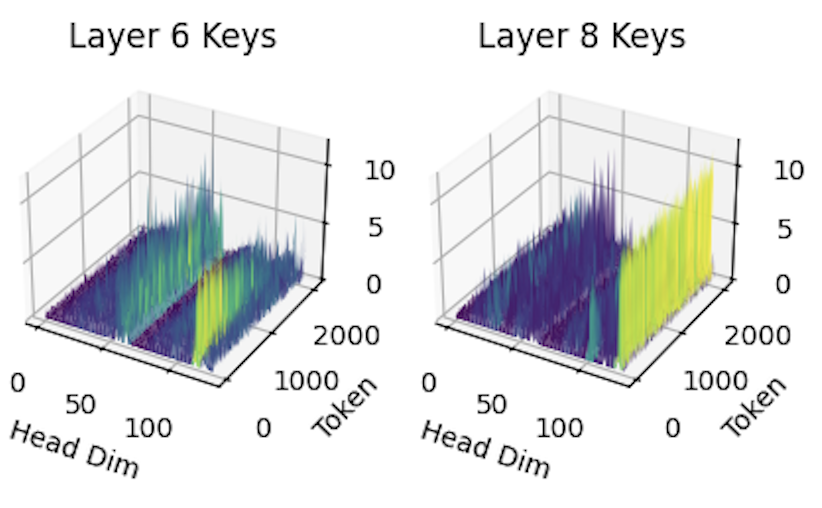}}
\end{subcaptionbox}
\hfill
\begin{subcaptionbox}{Magnitude distribution of Value cache\label{fig:sub2}}[0.45\textwidth]
  {\includegraphics[width=\linewidth]{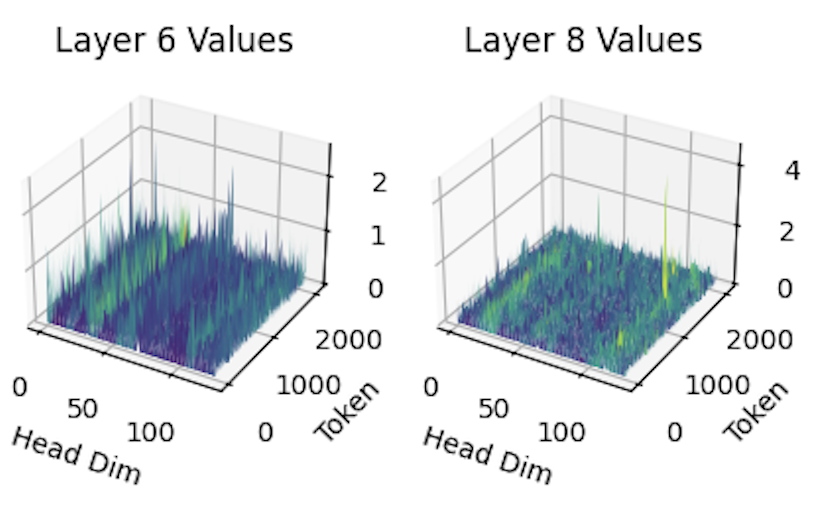}}
\end{subcaptionbox}
\caption{Visualization of the KV cache in LLaMA-2 7B. Color intensity indicates element magnitude. The figure was generated using the visualization code from KIVI~\cite{kivi}.} 
\label{fig:kv_visualization}
\end{figure}

In deciding the pruning direction, we build on top of the observation of KIVI~\cite{kivi}, that Key cache exhibits distinct channel-wise outliers, where "channel" refers to the head dimension (Figure~\ref{fig:sub1}). This leads us to focus on per-token pruning for key cache, as it can effectively capture the elements in the outlier channel. 

Based on the same observation to perform structured pruning of individual channels, ThinK~\cite{think} incorporates output-awareness by using a per-channel score of the accumulation of last 32 query, multiplied by each channel. To this end we compare the accuracy of ThinK~\cite{think}, per-token magnitude-based unstructured pruning, and output-aware unstructured pruning of our design. 

\begin{wrapfigure}{l}{0.5\textwidth}  
    \centering
    \includegraphics[width=0.50\textwidth]{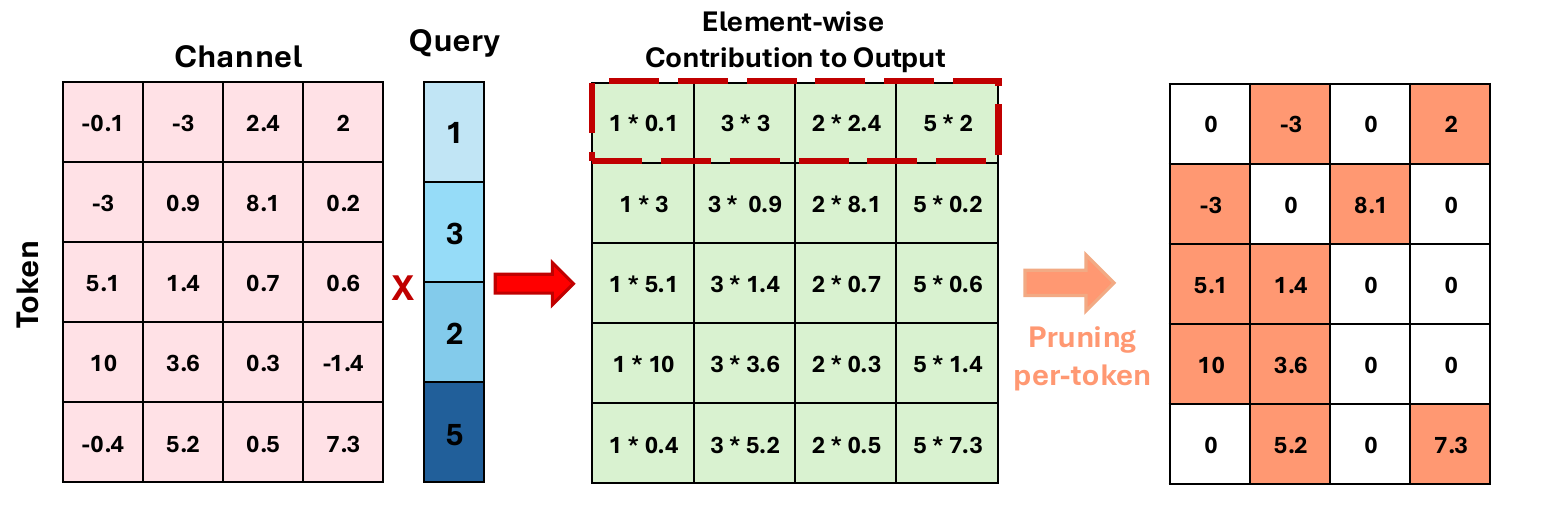}
    \caption{Per-token, output-aware pruning of Key cache}
    \label{fig:key_opa}
\end{wrapfigure}

Figure~\ref{fig:key_opa} elaborates the per-token output-aware unstructured pruning score of Key cache. The element-wise $L_1$ accumulation of the current and next 31 Query vector (blue) is multiplied element-wise across each token's key vector (pink) to derive the pruning score (green). The absolute value of the score element in the corresponding position of each Key cache element is used to decide the elements to be pruned within a token's Key vector. 
%Magnitude-based sorting is performed o is used to  with with elements to prune per-token is decided to reach the target sparsity.
In other words, we formulate the per-token output-aware unstructured pruning score $ S $ of a Key cache $ K $ to be:

\[
S = |K| \odot broadcast \left( \sum_{t=T}^{T+31} |Q_t| \right), \quad \text{where } Q_t \text{ is the query at time } t
\]

For Group Query Attention (GQA)~\cite{ainslie2023gqa}, where multiple queries correspond to a KV cache pair, we sum the pruning score of all queries mapped to each KV cache. 

\begin{table}[h]
\centering
\caption{Comparison of ThinK~\cite{think} structured pruning, per-token magnitude-based unstructured pruning, and per-token output-aware unstructured pruning on LongBench~\cite{bai2024longbench} with Llama-3-8B-Instruct Key cache. $K_s$ denotes Key cache sparsity.}
\label{tab:k_compare}
\vspace{0.5em}
\resizebox{\textwidth}{!}{%
\begin{tabular}{|c|c|ccc|ccc|}
\hline
%\multirow{2}{*}{Dataset} & \textbf{Dense} & \multicolumn{2}{c|}{K = 0.5} & \multicolumn{2}{c|}{K = 0.7} \\
\multirow{3}{*}{\centering\textbf{Task}} & \multirow{3}{*}{\centering\textbf{Dense}} 
& \multicolumn{3}{c|}{$K_s$ = 0.5} & \multicolumn{3}{c|}{$K_s$ = 0.7}  \\
\cline{3-8}
%& & ThinK (Structured) & Mustafar (Unstructured) & ThinK (Structured) & Mustafar (Unstructured) \\
& & \makecell{ThinK \\ (Structured)} & \makecell{Unstructured \\ Output-aware} & \makecell{Unstructured \\ Magnitude} & \makecell{ThinK \\ (Structured)} & \makecell{Unstructured \\ Output-aware} & \makecell{Unstructured \\ Magnitude} \\
\hline
Average       & 43.19   & 38.53 & \textbf{43.23} & 42.84     & 26.55  & \textbf{42.13} & 41.55 \\
\hline
SingleDoc QA  & 36.66   & 35.61 & 36.57 & \textbf{36.90}     & 25.26  & \textbf{35.78} & 35.53 \\
MultiDoc QA   & 36.09   & 34.99 & \textbf{35.92} & 35.77     & 29.75  & \textbf{35.55} & 35.40 \\
Summarization & 26.75   & 24.96 & \textbf{26.87} & 26.45     & 17.70  & 25.16 & \textbf{25.18} \\
Few-shot      & 68.96   & 66.54 & \textbf{68.82} & 68.75     & 44.88  & 67.22 & \textbf{67.84} \\
Synthetic     & 37.25   & 35.50 & \textbf{37.00} & 36.75     & 16.86  & \textbf{35.25} & 35.00 \\
Code          & 55.58   & 29.56 & \textbf{56.61} & 54.14     & 19.15  & \textbf{56.19} & 51.47 \\
\hline
\end{tabular}
}
\end{table}

In Table~\ref{tab:k_compare}, we compare Llama-3-8B-Instruct accuracy of different pruning methods on LongBench~\cite{bai2024longbench}. For structured pruning, we see that even at a moderate sparsity, model accuracy retention is dismal compared to pruning to an unstructured sparsity pattern. 
Notably, unstructured pruning is capable of outperforming structured pruning even without the memory footprint of pruning scores involved with output-awareness. Applying output-awareness to unstructured pruning results in a slight improvement in the LongBench total average score, while individual task performance is mixed with each method outperforming the other on different tasks.

\begin{tcolorbox}[colback=blue!5!white, colframe=blue!5!white]
\textbf{Key Cache Verdict:} While the existence of outlier channels with exceptionally high magnitudes show promise for per-channel structured pruning, unstructured sparsity achieves higher accuracy at greater sparsity levels, even without output-awareness.
\end{tcolorbox}

\subsection{Pruning Value Cache}

As shown in Figure~\ref{fig:sub2}, Value cache exhibits more uniform distribution of activations, making it challenging to apply the same channel-wise pruning without incurring substantial degradation in model accuracy. This difficulty has led recent Value cache pruning approaches to be more susceptible to accuracy degradation. 

\begin{wrapfigure}{r}{0.5\textwidth}  
    \centering
    \includegraphics[width=0.50\textwidth]{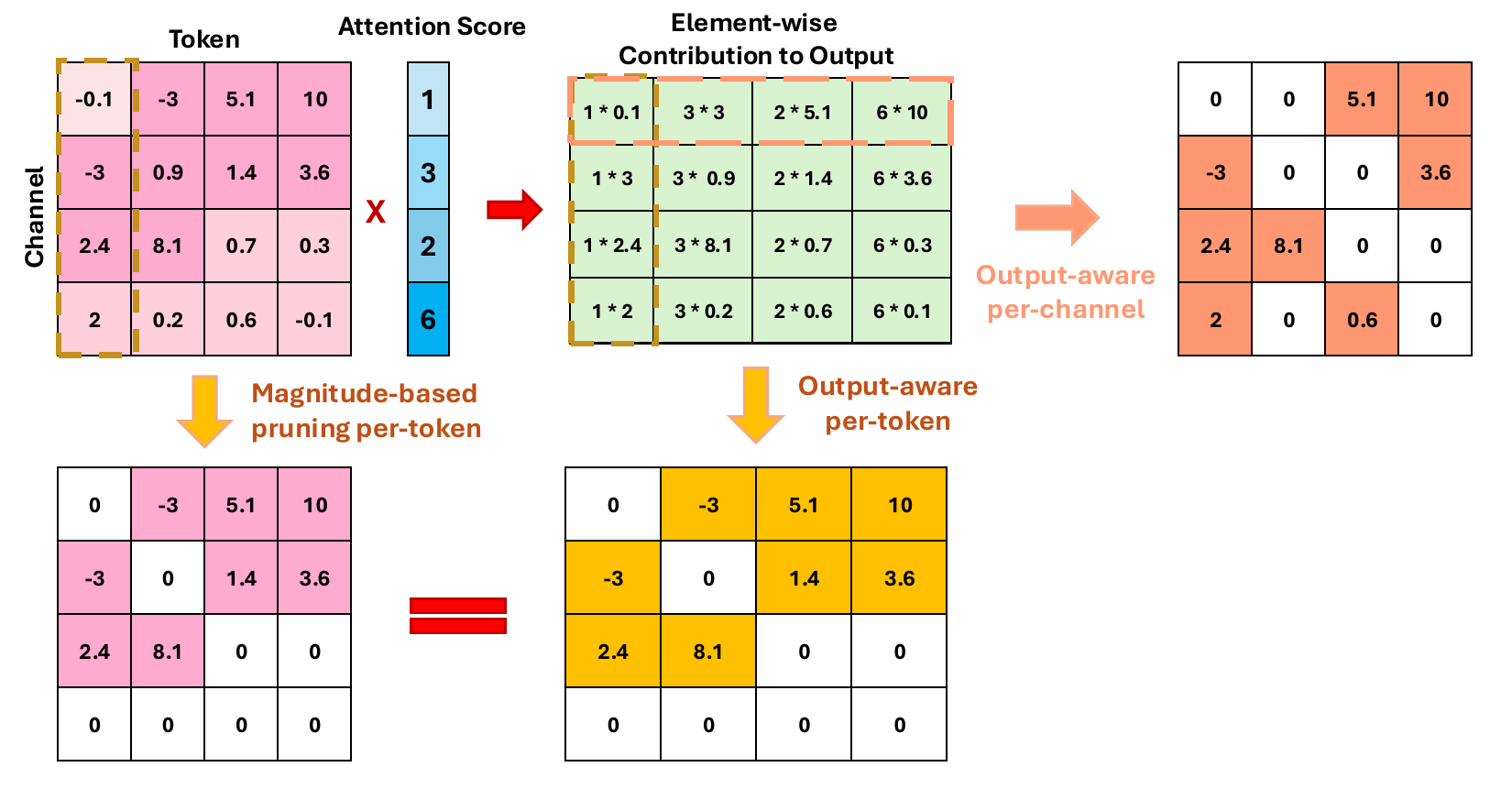}
    \caption{Output-aware per-channel (red) and magnitude-based per-token (pink) pruning of Value cache. Magnitude-based per-token pruning is equal to output-aware per-token pruning (yellow).}
    \label{fig:value_opa}
\end{wrapfigure}

With no discernible outliers in certain direction, we explore all possible combinations of (pruning direction, magnitude/output-aware) pairs. However, we are able to rule out per-token output-aware pruning, as the attention formulation $Attention Score \times Value$ involves a multiply-and-accumulate operation along the token dimension.
%However, we are able to rule out per-token output-aware pruning, due to the fact that in attention formulation of $Attention Score \times Value$, multiply-and-accumulate is performed on the token-dimension. 
As seen in Figure~\ref{fig:value_opa}, every element of a token's Value cache is multiplied by the same element of the attention score, with each element's impact on the output proportionate to the magnitude of each value. \textbf{That is, for Value cache pruning, per-token magnitude-based pruning is already output-aware}. For per-channel pruning, we prune each channel to the target sparsity in groups of 32 tokens, for compatibility with the local window size. 
For per-channel output-aware pruning, we accumulate the current and subsequent 31 attention score $\alpha$ of each token, which is then element-wise multiplied to the corresponding Value Cache ($V$) element. The following formula describes the pruning score $S$ of per-channel output-aware pruning: 
\[
S = |V| \odot broadcast\left( \sum_{t=T}^{T+31} |\alpha_t| \right), \quad \text{where } \alpha_t \text{ is the attention score at time } t
\]

\begin{table}[h!]
\centering
\caption{Comparison of ThinK~\cite{think} structured pruning, per-channel magnitude-based unstructured pruning, per-channel output-aware unstructured pruning, and per-token magnitude-based pruning on LongBench~\cite{bai2024longbench} with Llama-3-8B-Instruct Value Cache. $V_s$ denotes Value cache sparsity.}
\label{tab:v_compare}
\vspace{0.5em}
\resizebox{\textwidth}{!}{%
\begin{tabular}{|c|c|cccc|cccc|}
\hline
\multirow{3}{*}{\centering\textbf{Task}} & \multirow{3}{*}{\centering\textbf{Dense}} 
& \multicolumn{4}{c|}{$V_s$ = 0.5} & \multicolumn{4}{c|}{$V_s$ = 0.7}  \\
\cline{3-10}
& & \makecell{ThinK \\ (Structured)} & \makecell{Magnitude \\ (Per-channel)} & \makecell{Output-aware \\ (Per-channel)} & \makecell{Magnitude \\ (Per-token)}  & \makecell{ThinK \\ (Structured)} & \makecell{Magnitude \\ (Per-channel)} & \makecell{Output-aware \\ (Per-channel)} & \makecell{Magnitude \\ (Per-token)} \\
\hline
Average       & 43.19  & 38.45 & 42.50 & 42.84 & \textbf{43.04}     & 30.60 & 41.69 & 42.67 & \textbf{42.78} \\
\hline
SingleDoc QA  & 36.66  & 34.92 & 36.56 & 36.24 & \textbf{36.75}     & 25.05 & 36.11 & 36.05 & \textbf{36.96} \\
MultiDoc QA   & 36.09  & 34.74 & 35.45 & 36.07 & \textbf{36.22}     & 23.90 & 35.11 & \textbf{36.20} & 35.82 \\
Summarization & 26.75  & 23.31 & 24.74 & 25.79 & \textbf{26.34}     & 20.41 & 22.72 & 24.75 & \textbf{25.19} \\
Few-shot      & 68.96  & 67.18 & 67.66 & 68.65 & \textbf{68.91}     & 60.16 & 67.39 & \textbf{68.23} & 68.08 \\
Synthetic     & 37.25  & 35.43 & \textbf{38.31} & 37.00 & 36.25     & 29.63 & \textbf{38.75} & 37.25 & 35.50 \\
Code          & 55.58  & 31.97 & 55.07 & 55.57 & \textbf{55.77}     & 20.85 & 52.65 & 56.17 & \textbf{57.62} \\
\hline
\end{tabular}
}
\end{table}

As shown in the Table~\ref{tab:v_compare}, we first see that applying structured pattern to Value cache pruning incurs significant accuracy degradation even in 50\% sparsity. This is concurrent with ThinK~\cite{think} findings, which points to 30\% sparsity as the upper-bound on acceptable accuracy. In contrast, per-token magnitude pruning is capable of preserving model accuracy even at 70\% sparsity.
For per-channel pruning, we see that incorporating output-awareness boasts model accuracy retention almost to the level of per-token pruning. However, we prefer per-token magnitude-based pruning for the following two reasons. 
First, output-aware per-channel value cache pruning requires access to the attention score which requires additional recomputation when used alongside FlashAttention~\cite{flashattention2}, where the full attention score matrix does not materialize in the global memory.
Second, per-token magnitude-based pruning allows smooth compatibility with orthogonal compression method token-wise eviction~\cite{infinigen, h2o}, where the retained token's KV cache can be pruned individually. We examine the accuracy of joint application in Section~\ref{sec:joint_apply}.

\begin{tcolorbox}[colback=blue!5!white, colframe=blue!5!white]
\textbf{Value Cache Verdict}: All unstructured pruning methods explored outperform structured pruning. Among unstructured pruning methods, token-wise pruning, which is inherently output-aware by matrix multiplication formulation, best preserves model accuracy even at high sparsity levels. While channel-wise pruning with output-awareness can achieve comparable accuracy, token-wise pruning offers advantages in both efficiency and modularity.
\end{tcolorbox}

With the two verdicts in Key and Value caches, on Table~\ref{tab:kv_compare} we finally validate the model accuracy retention of per-token magnitude-based pruning with both Key and Value caches pruned. Not only can Value cache be pruned to high sparsity with unstructured sparsity, but both KV cache can be pruned to 70\% sparsity while showing similar or better accuracy than Key-only 50\% structured pruning of ThinK~\cite{think}. In Appendix~\ref{appendix:llama2}, methodology of this section is applied on Llama-2 7B to reinforce the effectiveness of per-token magnitude-based KV cache pruning. 

\begin{table}[h!]
\centering
\caption{Longbench Score of Llama-3-8B-Instruct and Mistral-7B-Instruct-v0.2 with KV Cache Per-Token Magnitude-based Pruning.}
\label{tab:kv_compare}
\vspace{0.5em}
\resizebox{0.70\linewidth}{!}{%
\begin{tabular}{|c|ccc|ccc|}
\hline
\multirow{3}{*}{Task} & \multicolumn{3}{c|}{Llama-3-8B-Instruct} & \multicolumn{3}{c|}{Mistral-7B-Instruct-v0.2} \\
\cline{2-7}
&  \multirow{1}{*}{\textbf{Dense}} & \makecell{$K_s$ = 0.5 \\ $V_s$ = 0.5} & \makecell{$K_s$ = 0.7 \\ $V_s$ = 0.7} & \multirow{1}{*}{\textbf{Dense}} & \makecell{$K_s$ = 0.5 \\ $V_s$ = 0.5} & \makecell{$K_s$ = 0.7 \\ $V_s$ = 0.7} \\
\hline
Average        & 43.19 & 42.65 & 40.96   & 42.65 & 42.30 & 40.95 \\
\hline
SingleDoc QA   & 36.66 & 36.67 & 35.28   & 36.21 & 36.22 & 36.08 \\
MultiDoc QA    & 36.09 & 36.23 & 35.11   & 29.93 & 30.42 & 29.40 \\
Summarization  & 26.75 & 26.05 & 23.57   & 28.10 & 27.77 & 26.72 \\
Few-shot       & 68.96 & 68.18 & 66.10   & 66.68 & 66.70 & 66.24 \\
Synthetic      & 37.25 & 36.00 & 34.13   & 44.85 & 41.92 & 36.13 \\
Code           & 55.58 & 54.50 & 53.49   & 54.98 & 54.83 & 53.84 \\
\hline
\end{tabular}
}
\end{table}

\section{Sparse Attention Kernel}
\label{sec:3}

Our findings establish that unstructured sparsity offers superior sparsity ratios over structured sparsity while preserving accuracy. In turn, a crucial contribution of Mustafar is to leverage this advantage to enable high compression efficiency while minimizing the latency overhead of runtime pruning and compression.
Prior compression methods such as quantization, structured pruning, and token eviction reduce matrix dimensions or element bitwidths. In terms of efficiency, speedup from the reduced size of dense matrix operands compensates for the additional latency introduced by compression (i.e. pruning score computation, quantization). In contrast, unstructured sparsity with no regular reduction in dimensions or element bitwidth demands a different approach. 

\begin{figure}[t]
\centering
\begin{subcaptionbox}{Coruscant~\cite{joo2025coruscant} bitmap-based sparse format\label{fig:sub_bmp}}[0.47\textwidth]
{\includegraphics[width=\linewidth]{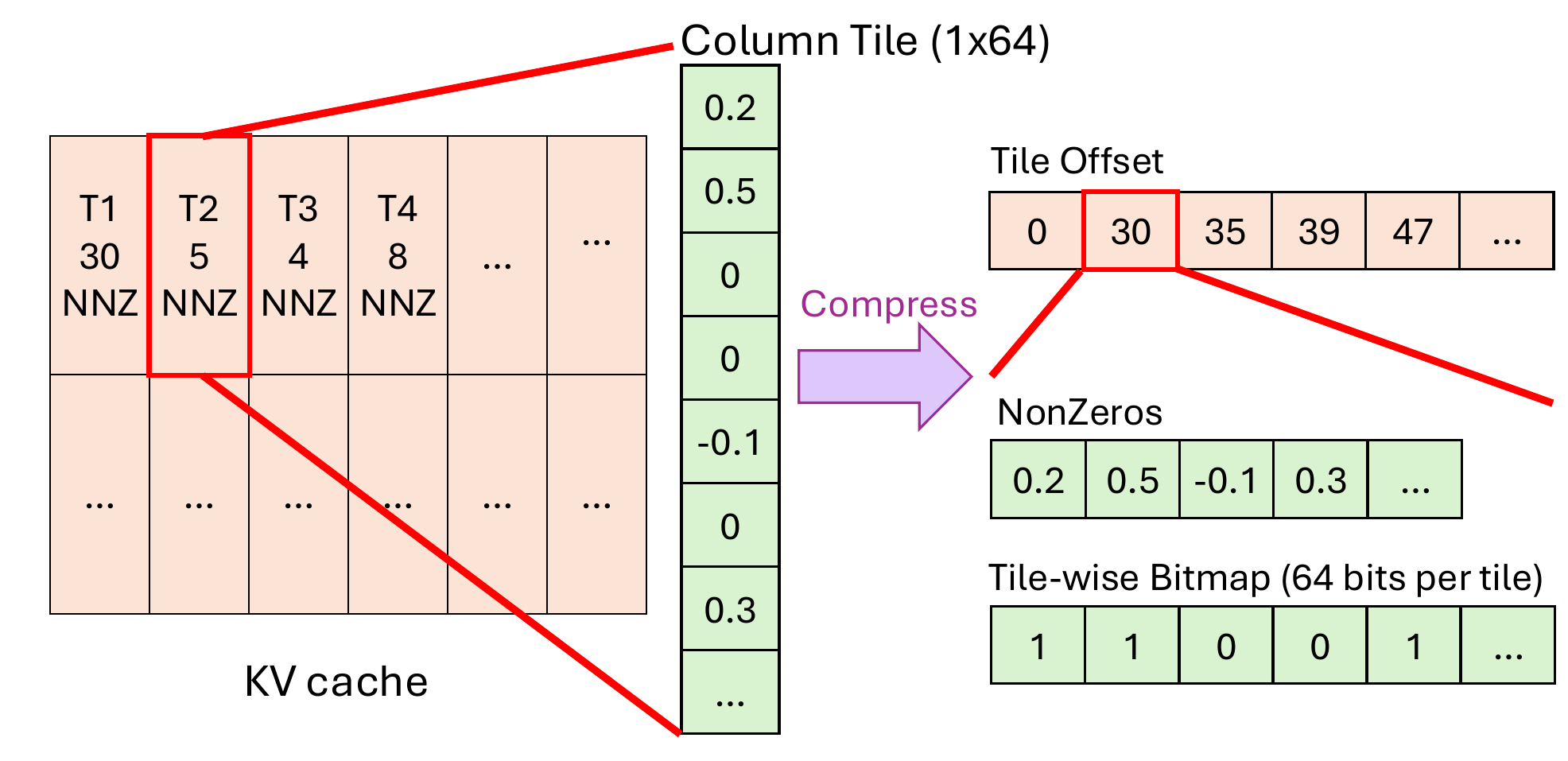}}
\end{subcaptionbox}
\hspace{0.03\textwidth}
\begin{subcaptionbox}{Mustafar attention kernel formulation\label{fig:sub_attn}}[0.47\textwidth]
{\includegraphics[width=\linewidth]{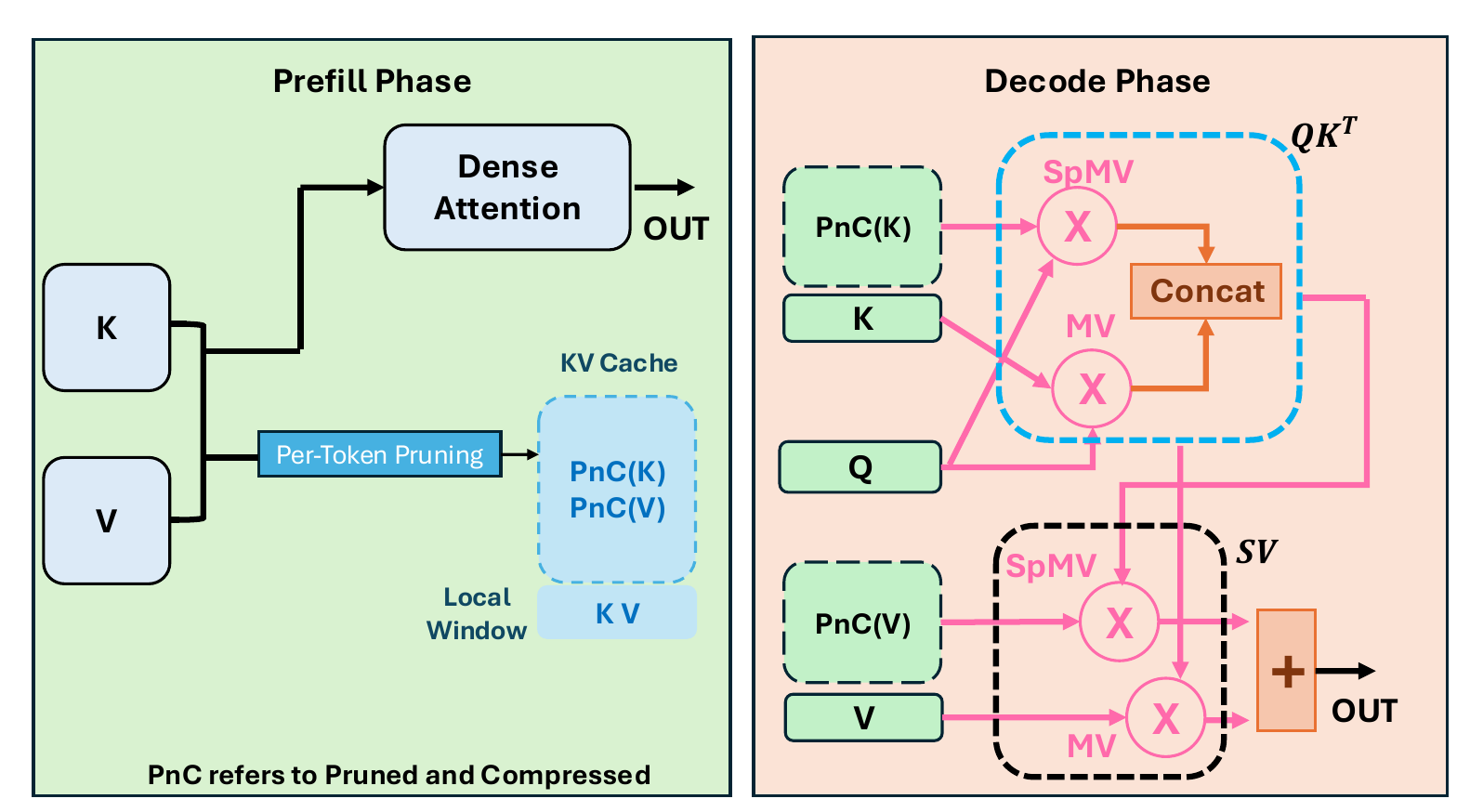}}
\end{subcaptionbox}
\caption{Overview of Mustafar sparse attention kernel. In (b), multi-head, softmax, and normalization are omitted for simplicity.} 
\label{fig:attention_kernel}
\end{figure}

Mustafar is motivated by the observation that attention operations in the autoregressive decode stage, the \textit{Query} $\times$ \textit{Key$^T$} and \textit{Attention Score} $\times$ \textit{Value} computations are batch (different heads) of matrix-vector products (MVs) that are significantly memory-bound on GPUs compared to the prefill stage. 
To exploit this property, we extend the bitmap-based sparse format of Coruscant~\cite{joo2025coruscant} as shown in Figure~\ref{fig:sub_bmp} to maximally compress the pruned KV cache. 
It consists of compressed tiles corresponding to a $1\times64$ column of the pruned cache. Per-tile bitmap of 64 bits is used to represent the position of non-zeros, and tile offset is used to address the correct position of each tile's starting non-zero.   
Pruning and compression are performed on-the-fly, with compression accelerated on GPU with a Triton kernel, and attention is computed directly on the compressed representation with a custom CUDA kernel that performs batch SpMV on the bitmap-based sparse format. Memory-bound decode-phase attention is accelerated by reducing the data movement from global memory to GPU Streaming Multiprocessors.

Figure~\ref{fig:sub_attn} and Algorithm~\ref{alg:sparse_attention_formula} presents the Mustafar sparse attention kernel. KV cache generated in prefill stage is pruned and compressed before the start of decode stage, therefore compatible with prefill FlashAttention~\cite{flashattention2}. KV cache generated in decode stage is kept as-is (dense) while it is within the local window, then pruned and compressed afterwards. This entails the attention computations in the decode stage to be reformulated into two parts: SpMV for compressed KV cache (line 2 and 5 of Algorithm~\ref{alg:sparse_attention_formula}) and dense MV for the KV cache within the local window (line 1 and 5 of Algorithm~\ref{alg:sparse_attention_formula}). 

\begin{algorithm}[h]
\caption{Decode Phase Attention with Dense Local and Compressed KV Caches}
\label{alg:sparse_attention_formula}
\begin{algorithmic}[1]
\Statex \textbf{Input:} 
Query $\mathbf{Q}_t \in \mathbb{R}^{d}$; 
Local KV cache $\mathbf{K}_L, \mathbf{V}_L \in \mathbb{R}^{d \times N_d}$, where $N_d$ is size of local window in tokens; 
Compressed KV cache $\mathbf{K}_C, \mathbf{V}_C \in \mathbb{R}^{d \times N_s}$, where $N_s$ is number of compressed tokens.

\vspace{2mm}
\Statex \textbf{Attention Score Computation}
\State $\mathbf{S}_L \in \mathbb{R}^{1 \times N_d} \gets \mathbf{Q}_t \mathbf{K}_L$ \hfill \textit{Dense local window attention score}
\State $\mathbf{S}_C  \in \mathbb{R}^{1 \times N_s} \gets \mathbf{Q}_t \mathbf{K}_C$ \hfill \textit{Sparse attention score over compressed KV cache}
\State $\mathbf{S}_t \in \mathbb{R}^{1 \times (N_s + N_d)} \gets \mathrm{softmax}\!\big(\mathrm{concat}(\mathbf{S}_C, \mathbf{S}_L)\big)$ 
      \hfill \textit{Full attention score}

\vspace{2mm}
\Statex \textbf{Output Computation}
\State $[\mathbf{S}_C, \mathbf{S}_L] \gets \mathrm{split}(\mathbf{S}_t; N_s, N_d)$ \hfill \textit{Partition attention score}
\State $\mathbf{O}_t \in \mathbb{R}^{d} \gets \mathbf{V}_C \mathbf{S}_C^\top + \mathbf{V}_L \mathbf{S}_L^\top$ 
      \hfill \textit{Final output vector}

\vspace{1mm}
\Statex \textbf{Return} $\mathbf{O}_t$
\end{algorithmic}
\end{algorithm}

%\paragraph{Let:}
%\begin{align*}
%&\text{Query vector at decoding step } t: && \mathbf{Q}_t \in \mathbb{R}^{d}, \\
%&\text{Key vectors within the local window: } && \mathbf{K}_L \in \mathbb{R}^{d \times N_d}, \text{ where } N_d \text{ is the size of local window in tokens.} \\
%&\text{Compressed Key vectors outside the local window: } && \mathbf{K}_C \in \mathbb{R}^{d \times N_s}, \text{ where } N_s \text{ is the number of compressed tokens.} \\
%&\text{Corresponding Value caches: } && \mathbf{V}_L \in \mathbb{R}^{d \times N_d}, \quad \mathbf{V}_C \in \mathbb{R}^{d \times N_s}.
%\end{align*}

%\paragraph{Attention score computation:}
%\begin{align*}
%&\text{Dense local window attention score: } && \mathbf{S}_L = \mathbf{Q}_t \cdot \mathbf{K}_L, \quad \mathbf{S}_L \in \mathbb{R}^{1 \times N_d}, \\
%&\text{Sparse attention score over pruned KV cache: } && \mathbf{S}_C = \mathbf{Q}_t \cdot \mathbf{K}_C, \quad \mathbf{S}_C \in \mathbb{R}^{1 \times N_s}, \\
%&\text{Full attention weight: } && \mathbf{S}_t = \mathrm{softmax}(\mathrm{concat}(\mathbf{S}_C, \mathbf{S}_L)), \quad \mathbf{S}_t \in \mathbb{R}^{1 \times (N_s + N_d)}.
%\end{align*}

%\paragraph{Final output computation:}
%\begin{align*}
%[\mathbf{S}_C, \mathbf{S}_L] &= \mathrm{split}(\mathbf{S}_t; N_s, N_d), \\
%\mathbf{O}_t &= \mathbf{V}_C \cdot \mathbf{S}_C^\top + \mathbf{V}_L \cdot \mathbf{S}_L^\top, \quad \mathbf{O}_t \in \mathbb{R}^{d}.
%\end{align*}

Mustafar SpMV kernel follows the load-as-compressed, compute-as-dense paradigm adopted by FlashLLM~\cite{flashllm}, SpInfer~\cite{fan2025spinfer}, and Coruscant~\cite{joo2025coruscant}, which target sparse matrix–dense matrix multiplication in LLM weight projection layers. The compressed KV cache is loaded from GPU global memory into registers in its compressed form, decompressed into shared memory, and then used for tile-wise dense computation. We evaluate the performance of the Mustafar attention kernel and quantify the runtime overhead of pruning and compression in Section~\ref{sec:kernel_eval}. We further detail the formulation of the SpMV kernel, as well as the management of the compressed KV cache in Appendix~\ref{sec:appendix_kernel}.

\section{Evaluation}
%also think about a nice way to point to the results on appendices. 
\textbf{Methodology}: We evaluate Mustafar on two aspects: Accuracy and Efficiency. 
For accuracy evaluation, we use tasks from LongBench~\cite{bai2024longbench} 
%and CoQA, TruthfulQA, and GSM8K from LM-Eval-Harness 
to test the accuracy retention of per-token magnitude-based pruning of KV cache. 
We evaluate on three models: Llama-2-7B~\cite{touvron2023llama2}, Llama-3-8B-Instruct~\cite{grattafiori2024llama3}, and Mistral-7B-Instruct-v0.2~\cite{jiang2023mistral7b}.
We also explore the impact of Mustafar when jointly used with orthogonal compression techniques, KV cache quantization KIVI~\cite{kivi} and token-wise eviction H2O~\cite{h2o}.
%This is what KIVI did,
%NIAH? 
%For accuracy evaluation, we select 8 representative testbenches from LongBench and 4 representative testbenches from LM-Eval-Harness. 
For efficiency evaluation, we evaluate the impact on KV cache compression and computational latency with Llama-2-7B and Llama-3-8B-Instruct. Efficiency evaluation is tested on NVIDIA RTX 6000ADA GPU and measured with NVIDIA Nsight Profiling Tool.  
Additionally, we provide accuracy evaluation on RULER~\cite{hsieh2024ruler} benchmark in Appendix~\ref{sec:appendix_ruler}, accuracy comparison of Mustafar's unstructured sparsity with 2:4 semi-structured sparsity in Appendix~\ref{sec:appendix_2_4}, and additional kernel throughput evaluation in Appendix~\ref{sec:appendix_kernel_eval}.

%\subsection{Accuracy Evaluation}
%We see that ideal Joint pruning of KV cache is when the target sparsity if at 50\% sparsity. While Key pruning alone achieves significant accuracy retention even in 70\% sparsity, jointly using with Value pruning degrades performance.
\subsection{LongBench Results}
Table~\ref{tab:longbench_total} shows the extended LongBench evaluation of Mustafar per-token magnitude-based pruning with comparison to dense model and ThinK~\cite{think}. Under the same Key cache sparsity, unstructured nature of Mustafar constantly achieves higher accuracy than structured sparsity on ThinK~\cite{think} across all tasks. A key advantage of unstructured pruning is its ability to effectively prune the Value cache with minimal accuracy degradation, which structured pruning has struggled to achieve. Even under high sparsity 70\% for both the Key and Value caches, unstructured pruning (yellow) consistently outperforms ThinK's Key-only 50\% structured pruning (pink) on LLaMA-3 8B and Mistral 7B, and achieves comparable accuracy on LLaMA-2-7B.
%\ToDoD[]{\textbf{In words, reemphasize why this is better than previous works.}}

%\ToDoD[]{add LM-EVAL and NIAH in there, and LongGenBench}

\begin{table}[ht]
\centering
\caption{Mustafar accuracy with Llama and Mistral on LongBench}
\label{tab:longbench_total}
\vspace{0.5em}
\resizebox{\textwidth}{!}{%
\begin{tabular}{c|ccc|ccc|ccc|ccc|cc|cc|c}
\toprule
%\makecell{\textbf{KV} \\ \textbf{Sparsity}} & \multicolumn{3}{c|}{\textbf{Single-Document QA}} & \multicolumn{3}{c|}{\textbf{Multi-Document QA}} & \multicolumn{3}{c|}{\textbf{Summarization}} & \multicolumn{3}{c|}{\textbf{Few-shot Learning}} & \multicolumn{2}{c|}{\textbf{Synthetic}} & \multicolumn{2}{c|}{\textbf{Code}} & \textbf{Avg.} \\
%\midrule  
 & \multicolumn{3}{c|}{\textbf{Single-Document QA}} & \multicolumn{3}{c|}{\textbf{Multi-Document QA}} & \multicolumn{3}{c|}{\textbf{Summarization}} & \multicolumn{3}{c|}{\textbf{Few-shot Learning}} & \multicolumn{2}{c|}{\textbf{Synthetic}} & \multicolumn{2}{c|}{\textbf{Code}} &  \\
\cline{2-17}  
\makecell[c]{\textbf{KV}\\\textbf{Sparsity}}
& \makecell{\rotatebox{60}{\textbf{NtrvQA}}}
& \makecell{\rotatebox{60}{\textbf{Qasper}}}
& \makecell{\rotatebox{60}{\textbf{MF-en}}}
& \makecell{\rotatebox{60}{\textbf{HotpotQA}}}
& \makecell{\rotatebox{60}{\textbf{2WikiMQA}}}
& \makecell{\rotatebox{60}{\textbf{Musique}}}
& \makecell{\rotatebox{60}{\textbf{GovReport}}}
& \makecell{\rotatebox{60}{\textbf{QMSum}}}
& \makecell{\rotatebox{60}{\textbf{MultiNews}}}
& \makecell{\rotatebox{60}{\textbf{TREC}}}
& \makecell{\rotatebox{60}{\textbf{TrivialQA}}}
& \makecell{\rotatebox{60}{\textbf{SAMSum}}}
& \makecell{\rotatebox{60}{\textbf{PCount}}}
& \makecell{\rotatebox{60}{\textbf{PRe}}}
& \makecell{\rotatebox{60}{\textbf{Lec}}}
& \makecell{\rotatebox{60}{\textbf{RBP}}}
% & \textbf{}  \\
& \makecell[c]{\textbf{Avg.}} \\   
\midrule
\multicolumn{18}{c}{\textbf{Llama-3 8B Instruct}} \\
Dense  & 23.39 & 43.38 & 43.22 & 46.39 & 38.66 & 23.22 & 29.91 & 22.56 & 27.77 & 74.50 & 90.28 & 42.11 & 4.50 & 70.00 & 57.11 & 54.05 & \textbf{43.19} \\
ThinK0.5  & 22.38 & 40.96 & 43.48 & 44.01 & 38.37 & 22.59 & 26.61 & 22.20 & 26.08 & 74.00 & 88.83 & 36.79 & 6.00 & 65.00 & 27.95 & 31.17 & \cellcolor{pink}\textbf{38.53} \\
K0.5 V0.0   & 23.40 & 43.68 & 43.63 & 46.00 & 38.60 & 22.72 & 29.39 & 22.33 & 27.64 & 74.50 & 90.66 & 41.09 & 5.00 & 68.50 & 55.89 & 52.39 & \textbf{42.84} \\
ThinK0.7  & 17.58 & 27.40 & 30.80 & 40.59 & 29.50 & 19.16 & 18.13 & 17.28 & 17.70 & 34.00 & 83.09 & 17.56 & 4.71 & 29.00 & 17.88 & 20.42 & \textbf{26.55} \\
K0.7 V0.0  & 22.91 & 42.36 & 41.33 & 45.53 & 38.50 & 22.16 & 26.63 & 21.90 & 27.00 & 73.00 & 90.83 & 39.68 & 4.50 & 65.50 & 51.94 & 50.99 & \textbf{41.55} \\
K0.0 V0.5 & 23.80 & 43.14 & 43.32 & 46.28 & 39.42 & 22.97 & 29.18 & 22.70 & 27.13 & 74.50 & 90.50 & 41.74 & 5.00 & 67.50 & 57.23 & 54.30 &\textbf{43.04} \\
K0.0 V0.7 & 24.19 & 42.78 & 43.92 & 45.82 & 39.11 & 22.53 & 26.92 & 22.52 & 26.12 & 74.00 & 90.36 & 39.88 & 5.50 & 65.50 & 59.18 & 56.05 & \textbf{42.77}   \\
K0.5 V0.5 & 23.40 & 46.63 & 42.98 & 46.28 & 39.27 & 23.13 & 28.29 & 22.78 & 27.07 & 74.00 & 90.58 & 39.97 & 5.00 & 67.00 & 55.54 & 53.46 & \textbf{42.65} \\
K0.7 V0.7 & 24.10 & 40.85 & 40.88 & 44.93 & 38.03 & 22.36 & 24.02 & 21.90 & 24.78 & 70.50 & 90.04 & 37.77 & 5.25 & 63.00 & 54.12 & 52.86 & \cellcolor{yellow}\textbf{40.96} \\
\midrule
\multicolumn{18}{c}{\textbf{Mistral-7B-Instruct-v0.2}} \\
Dense  & 26.76 & 32.51 & 49.36 & 43.49 & 27.48 & 18.81 & 32.95 & 24.36 & 27.00 & 71.00 & 86.23 & 42.80 & 2.89 & 86.81 & 55.89 & 54.07 & \textbf{42.65} \\
ThinK0.5  & 24.03 & 26.79 & 46.42 & 38.70 & 24.93 & 15.73 & 32.72 & 24.65 & 27.14 & 71.00 & 85.80 & 41.68 & 2.20 & 73.67 & 48.83 & 47.09 & \cellcolor{pink}\textbf{39.46} \\
K0.5 V0.0   & 26.38 & 33.08 & 49.20 & 43.90 & 28.57 & 18.65 & 32.47 & 24.21 & 27.05 & 71.00 & 86.28 & 42.66 & 3.00 & 84.23 & 55.72 & 54.16 & \textbf{42.56} \\
ThinK0.7  & 19.25 & 21.33 & 36.48 & 27.96 & 20.34 & 14.08 & 29.32 & 22.23 & 25.64 & 70.50 & 78.99 & 29.66 & 2.92 & 54.42 & 34.28 & 31.68 & \textbf{32.44} \\
K0.7 V0.0  & 27.02 & 34.37 & 49.26 & 43.77 & 26.37 & 17.45 & 32.05 & 24.09 & 27.43 & 71.00 & 87.19 & 42.30 & 4.65 & 77.24 & 54.26 & 53.06 & \textbf{41.97} \\
K0.0 V0.5 & 26.29 & 32.54 & 49.01 & 43.99 & 28.02 & 19.28 & 32.07 & 23.74 & 26.98 & 71.00 & 86.56 & 42.79 & 2.71 & 81.77 & 55.14 & 54.16 & \textbf{42.25} \\
K0.0 V0.7 & 26.83 & 31.66 & 49.24 & 44.15 & 27.40 & 18.36 & 30.58 & 23.80 & 26.63 & 71.00 & 86.82 & 42.02 & 3.77 & 76.32 & 55.58 & 54.16 & \textbf{41.77} \\
K0.5 V0.5 & 26.90 & 32.99 & 48.76 & 43.90 & 28.90 & 18.45 & 32.24 & 24.09 & 26.99 & 71.00 & 86.68 & 42.41 & 3.20 & 80.64 & 55.51 & 54.15 & \textbf{42.30} \\
K0.7 V0.7 & 27.11 & 32.23 & 48.90 & 43.63 & 27.12 & 17.43 & 29.38 & 23.99 & 26.79 & 71.00 & 86.59 & 41.14 & 4.69 & 67.57 & 54.86 & 52.82 & \cellcolor{yellow}\textbf{40.95} \\
\midrule
\multicolumn{18}{c}{\textbf{Llama-2 7B}} \\
Dense  & 15.04 & 9.66 & 21.88 & 7.69 & 9.95 & 3.66 & 17.26 & 21.29 & 3.5 & 66.00 & 87.72 & 41.66 & 1.70 & 6.64 & 66.66 & 59.82 & \textbf{27.51} \\
ThinK0.5  & 15.57 & 9.96 & 23.31 & 6.50 & 9.62 & 2.77 & 1.84 & 20.16 & 0.38 & 66.00 & 85.53 & 41.48 & 2.04 & 2.79 & 64.77 & 58.36 & \cellcolor{pink}\textbf{25.69} \\
K0.5 V0.0   & 14.79 & 9.65 & 21.67 & 7.48 & 10.10 & 4.11 & 17.24 & 20.84 & 3.64 & 66.00 & 87.72 & 41.26 & 1.38 & 6.42 & 67.15 & 59.89 & \textbf{27.46} \\
ThinK0.7  & 13.76 & 8.16 & 20.59 & 4.53 & 6.24 & 2.23 & 12.96 & 14.88 & 0.01 & 66.00 & 80.48 & 26.95 & 1.77 & 6.93 & 40.73 & 38.97 & \textbf{21.57} \\
K0.7 V0.0  & 14.57 & 8.18 & 20.55 & 6.64 & 9.95 & 3.28 & 13.80 & 20.25 & 0.88 & 66.00 & 86.64 & 38.32 & 2.12 & 4.04 & 64.86 & 58.59 & \textbf{26.17} \\
K0.0 V0.5 & 15.71 & 10.02 & 21.12 & 7.38 & 9.64 & 3.75 & 16.86 & 21.37 & 2.38 & 66.00 & 87.72 & 41.04 & 1.65 & 6.75 & 66.79 & 60.09 & \textbf{27.40} \\
K0.0 V0.7 & 15.57 & 8.98 & 20.97 & 7.33 & 10.14 & 3.82 & 15.40 & 20.77 & 1.83 & 66.00 & 87.72 & 40.69 & 1.40 & 6.50 & 66.12 & 59.57 & \textbf{27.05} \\
K0.5 V0.5 & 15.49 & 9.17 & 20.97 & 7.51 & 10.04 & 3.78 & 16.46 & 21.02 & 3.36 & 66.00 & 87.72 & 40.81 & 1.22 & 5.88 & 66.78 & 59.53 & \textbf{27.23} \\
K0.7 V0.7 & 13.76 & 7.83 & 19.27 & 6.57 & 10.26 & 3.51 & 8.70 & 20.04 & 0.47 & 64.50 & 86.89 & 36.37 & 1.64 & 3.62 & 63.95 & 56.75 & \cellcolor{yellow}\textbf{25.26} \\
\bottomrule
\end{tabular}%
}
\end{table}

%\newpage
\subsection{Joint Application with Orthogonal KV Cache Compression Techniques}
\label{sec:joint_apply}
Mustafar’s per-token pruning enables seamless integration with orthogonal KV cache compression techniques. We evaluate its effectiveness when combined with token eviction from H2O~\cite{h2o} and KV cache quantization from KIVI~\cite{kivi}, using a representative subset of LongBench tasks from each category. H2O application is conducted with Llama-2 7B and KIVI application is conducted with LLaMA-3-8B-Instruct. 

\subsubsection{Joint Application with Token Eviction}
H2O~\cite{h2o} retains a fixed budget of recent tokens and critical heavy-hitter tokens. Applying Mustafar to H2O, we retain the same scheme of pruning the KV cache of tokens that exit the local dense window. We configure 10\% of KV cache budget each to recent tokens and heavy-hitter tokens. Jointly applied, all heavy-hitter tokens and a part of recent tokens is kept as pruned and compressed. 
%Table~\ref{tab:joint_evict} presents the performance of Mustafar and H2O applied together. 
%Accuracy difference from Dense H2O method to pruned H2O is similar to that of full LLMs in Table~\ref{tab:longbench_total}. 
%Similar to previous results on full LLMs shown on Table~\ref{tab:k_compare}, we
In Table~\ref{tab:joint_evict}, we validate the efficacy of Mustafar's accuracy retention when jointly applied with token eviction, as we see that 50\% sparsity in both KV cache retains the dense accuracy with some degradation when pruned to 70\% sparsity. 

\begin{table}[ht]
\centering
\caption{LongBench evaluation of Mustafar-H2O joint application on Llama-2-7B}
\label{tab:joint_evict}
\vspace{0.5em}
\resizebox{0.8\linewidth}{!}{%
%\begin{tabular}{|p{2.0cm}|p{2.0cm}|p{2.0cm}|p{2.0cm}|p{2.0cm}|p{2.0cm}|p{2.0cm}|}
\begin{tabular}{c|c|c|c|c|c|c}
\toprule
\textbf{} & \textbf{Single-Doc QA} & \textbf{Multi-Doc QA} & \textbf{Summarization} & \textbf{Few-shot Learning} & \textbf{Synthetic} & \textbf{Code} \\
\textbf{} & \textbf{NtrvQA} & \textbf{HotpotQA} & \textbf{GovReport} & \textbf{TREC} & \textbf{Pcount} & \textbf{Lcc} \\
\midrule
%\multicolumn{7}{c}{\textbf{Llama-2 7B}} \\
\textbf{Full KV cache} & 15.04 & 7.69 & 17.26 & 66.00 & 1.7 & 66.66 \\
\midrule
\multicolumn{7}{c}{\textbf{H2O 20\% KV Budget}} \\
%\midrule
Dense          & 12.26 & 8.35 & 7.76 & 64.00 & 1.43 & 64.20 \\
K0.5 V0.0     & 11.83 & 8.47 & 7.65 & 64.00 & 2.08 & 64.72 \\
K0.7 V0.0     & 11.39 & 8.46 & 6.34 & 64.00 & 1.69 & 63.92 \\
K0.0 V0.5     & 12.17 & 8.39 & 6.84 & 64.00 & 1.38 & 64.83 \\
K0.0 V0.7     & 12.39 & 7.79 & 5.81 & 64.00 & 0.76 & 64.88 \\
K0.5 V0.5     & 12.07 & 8.16 & 7.61 & 64.00 & 2.05 & 65.15 \\
K0.7 V0.7     & 12.20 & 8.18 & 5.22 & 64.00 & 1.65 & 63.73 \\
\bottomrule
\end{tabular}%
}
\end{table}

\newpage
\subsubsection{Joint Application with Quantization} 

KIVI~\cite{kivi} applies a per-channel quantization of Key cache and per-token quantization of Value cache. Following findings of Harma et al.~\cite{harma2025interplay}, we first prune each token's KV cache before quantization is performed. However, we note that current Mustafar sparse attention kernel implementation does not support low-bit precision. Therefore, the accuracy measurement was performed on a sparse quantized KV cache.
Table~\ref{tab:joint_quant} shows the performance of Mustafar and KIVI applied together. Similar to joint application with H2O, we see that model accuracy is retained across the tasks for 50\% on Key cache, Value cache, as well as both Key and Value caches. We observe a decrease in accuracy at 70\% pruning, with Summarization task seeing the most significant drop. However, other tasks, such as Single-Document QA maintain the same performance as naive 16-bit model, suggesting the potential for applying varying degrees of compression tailored to specific tasks. 

\begin{table}[ht]
\centering
\caption{LongBench evaluation of Mustafar-KIVI joint application on Llama-3-8B-Instruct}
\label{tab:joint_quant}
\vspace{0.5em}
\resizebox{0.8\linewidth}{!}{%
%\begin{tabular}{|p{2.0cm}|p{2.0cm}|p{2.0cm}|p{2.0cm}|p{2.0cm}|p{2.0cm}|p{2.0cm}|}
\begin{tabular}{c|c|c|c|c|c|c}
\toprule
\textbf{} & \textbf{Single-Doc QA} & \textbf{Multi-Doc QA} & \textbf{Summarization} & \textbf{Few-shot Learning} & \textbf{Synthetic} & \textbf{Code} \\
\textbf{} & \textbf{NtrvQA} & \textbf{HotpotQA} & \textbf{GovReport} & \textbf{TREC} & \textbf{Pcount} & \textbf{Lcc} \\
\midrule
%\multicolumn{7}{c}{\textbf{Llama-2 7B}} \\
\textbf{Naive 16-bit} & 23.39 & 46.39 & 29.91 & 74.50 & 4.50 & 57.11 \\
\midrule
\multicolumn{7}{c}{\textbf{KIVI 4-bit}} \\
%\midrule
Dense         & 23.60 & 46.39 & 29.84 & 74.50 & 5.00 & 57.35 \\
K0.5 V0.0    & 23.46 & 46.21 & 28.90 & 74.50 & 5.50 & 56.05 \\
K0.7 V0.0    & 23.35 & 45.40 & 26.46 & 73.50 & 4.83 & 52.41 \\
K0.0 V0.5    & 23.68 & 46.39 & 29.10 & 74.50 & 5.50 & 58.30 \\
K0.0 V0.7    & 24.10 & 45.66 & 27.21 & 74.00 & 5.50 & 59.30 \\
K0.5 V0.5    & 23.22 & 46.06 & 28.18 & 74.00 & 6.00 & 56.04 \\
K0.7 V0.7    & 23.74 & 45.50 & 23.57 & 70.50 & 6.25 & 54.12 \\
\midrule
\multicolumn{7}{c}{\textbf{KIVI 2-bit}} \\
%\midrule
Dense         & 23.33 & 45.47 & 29.69 & 74.50 & 6.50 & 50.38 \\
K0.5 V0.0    & 22.86 & 45.29 & 29.39 & 74.00 & 5.50 & 49.92 \\
K0.7 V0.0    & 22.88 & 44.60 & 26.91 & 73.00 & 4.50 & 43.84 \\
K0.0 V0.5    & 23.65 & 45.67 & 29.05 & 74.00 & 5.50 & 51.94 \\
K0.0 V0.7    & 23.68 & 45.47 & 27.57 & 74.00 & 5.50 & 52.90 \\
K0.5 V0.5    & 22.46 & 45.47 & 28.61 & 74.00 & 4.50 & 48.76 \\
K0.7 V0.7    & 22.72 & 45.18 & 23.84 & 71.00 & 5.12 & 45.68 \\
\bottomrule
\end{tabular}%
}
\end{table}

\subsection{Efficiency Evaluation}
\label{sec:kernel_eval}

\begin{figure}[t]
\centering
\begin{subcaptionbox}{Normalized kernel latency breakdown.\label{fig:sub_latency_eval}}[0.47\textwidth]
{\includegraphics[width=\linewidth]{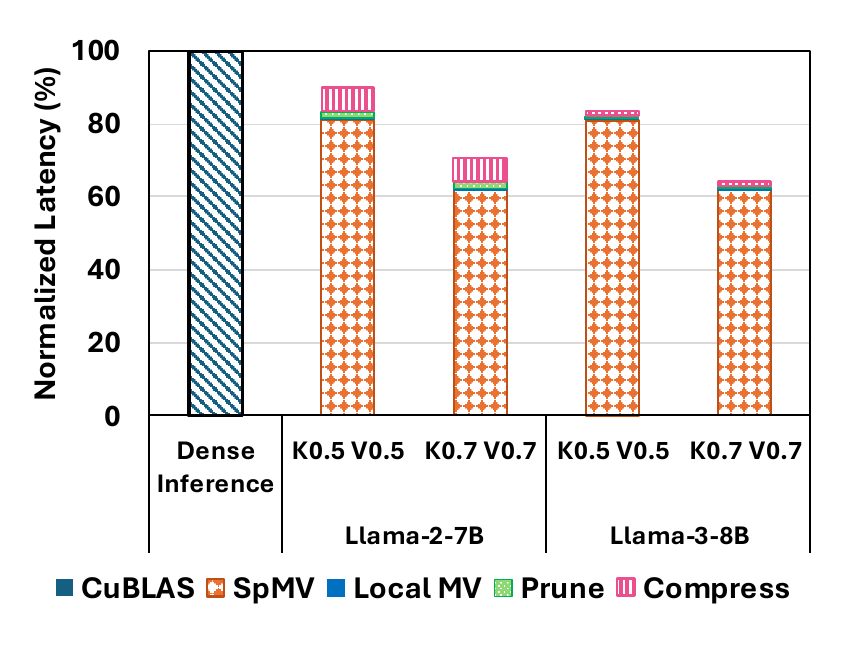}}
\end{subcaptionbox}
\hspace{0.03\textwidth}
\begin{subcaptionbox}{Compression ratio-accuracy comparison of Mustafar and ThinK.\label{fig:sub_compression_eval}}[0.47\textwidth]
{\includegraphics[width=\linewidth]{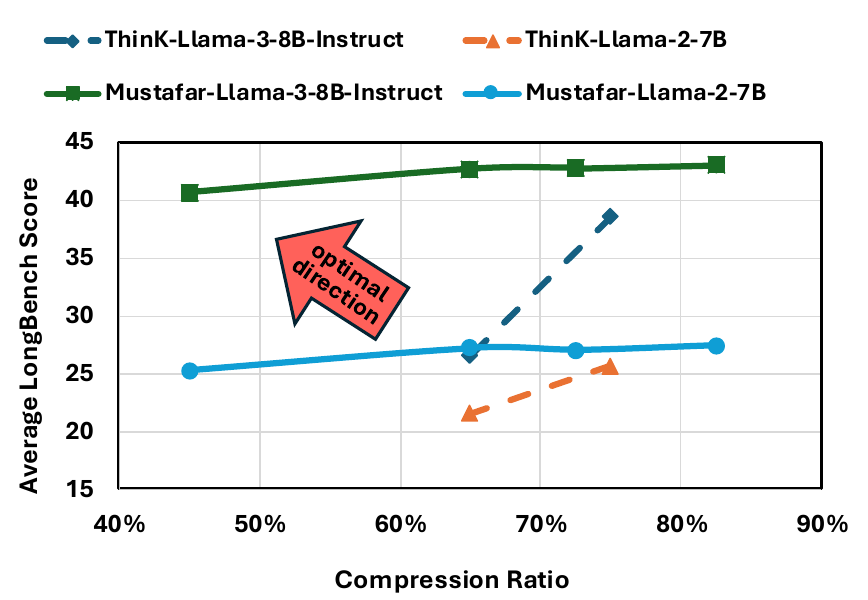}}
\end{subcaptionbox}
\caption{Efficiency evaluation of Mustafar. In (b), compression ratio refers to percentage of compressed size compared to dense KV cache.} 
\label{fig:attention_kernel}
\end{figure}

A crucial aspect of Mustafar is to ensure that the exploitation of sparsity for compressing the KV cache does not deter the inference latency. Mustafar compensates the overhead of runtime pruning and compression by achieving speedup in the memory-bound SpMV. Figure~\ref{fig:sub_latency_eval} compares the normalized latency of dense batched MV of cuBLAS with the components of Mustafar sparse attention kernel (Figure~\ref{fig:sub_attn}): batched SpMV, dense batched MV of local window, runtime pruning, and compression, for input sequence length 2048 for Llama-2 and 4096 for Llama-3 and generation length 1024. In the multi-head attention of Llama-2-7B, pruning introduces 1.84\%, compression introduces 6.25\%, and MV of local window introduces 0.62\% of the cuBLAS execution time in dense inference. In both 50\% and 70\%, the speedup gained from SpMV kernel more than compensates for the introduced overheads. In 50\% sparsity, SpMV takes 81.07\% of cuBLAS execution time and for 70\% sparsity, SpMV takes 61.87\% of cuBLAS execution time. In Grouped-Query Attention of Llama-3-8B, where there is reduced set of KV cache, compression and pruning overhead reduce down to 1.47\% and 0.47\% of cuBLAS execution time respectively. 

Figure~\ref{fig:sub_compression_eval} compares the KV cache compression ratio (\% of size in memory compared to dense KV cache) of Mustafar and ThinK along with the LongBench average score achieved with Llama-2-7B and Llama-3-8B-Instruct. In this plot, the red arrow points to the optimal direction, where a model achieves higher LongBench score while achieving high compression of the KV cache. For ThinK~\cite{think} which prune only Key cache, 50\% sparsity leads to 75\% compression ratio to dense KV cache, and 70\% Key cache sparsity leads to 65\% compression ratio. In the case of Mustafar where both Key and Value Cache can be pruned, KV cache 50\% sparsity leads to 65\% compression ratio. The reason behind 15\% additional memory footprint is due to the tile offset overhead as shown in Figure~\ref{fig:sub_bmp} and the multiples-of-8 padding enforced to coalesce memory access in GPU. KV cache 70\% sparsity leads to 45\% compression ratio, 50\% sparsity to either Key or Value cache leads to 83\% compression ratio, and single-cache 70\% sparsity leads to 72.5\% compression ratio. Overall, we see that Mustafar is able to achieve better accuracy given the compression ratio, with the compression ratio-accuracy curve closer to the optimal direction than ThinK. 

\begin{figure}[b]
\centering
\begin{subcaptionbox}{Llama-2 7B Throughput.\label{fig:sub_llama2_tput}}[0.47\textwidth]
{\includegraphics[width=\linewidth]{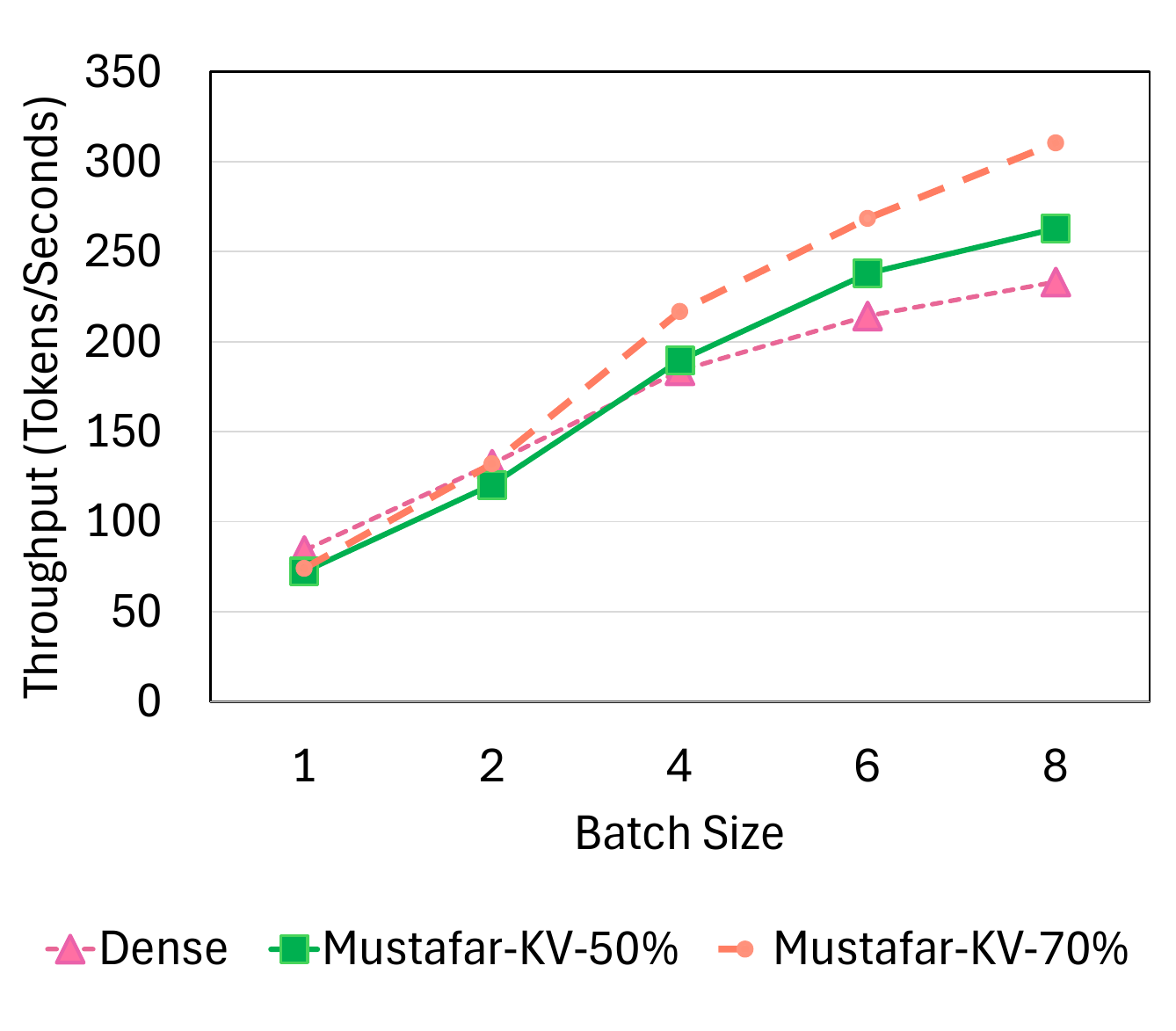}}
\end{subcaptionbox}
\hspace{0.03\textwidth}
\begin{subcaptionbox}{Llama-3 8B Instruct Throughput.\label{fig:sub_llama3_tput}}[0.47\textwidth]
{\includegraphics[width=\linewidth]{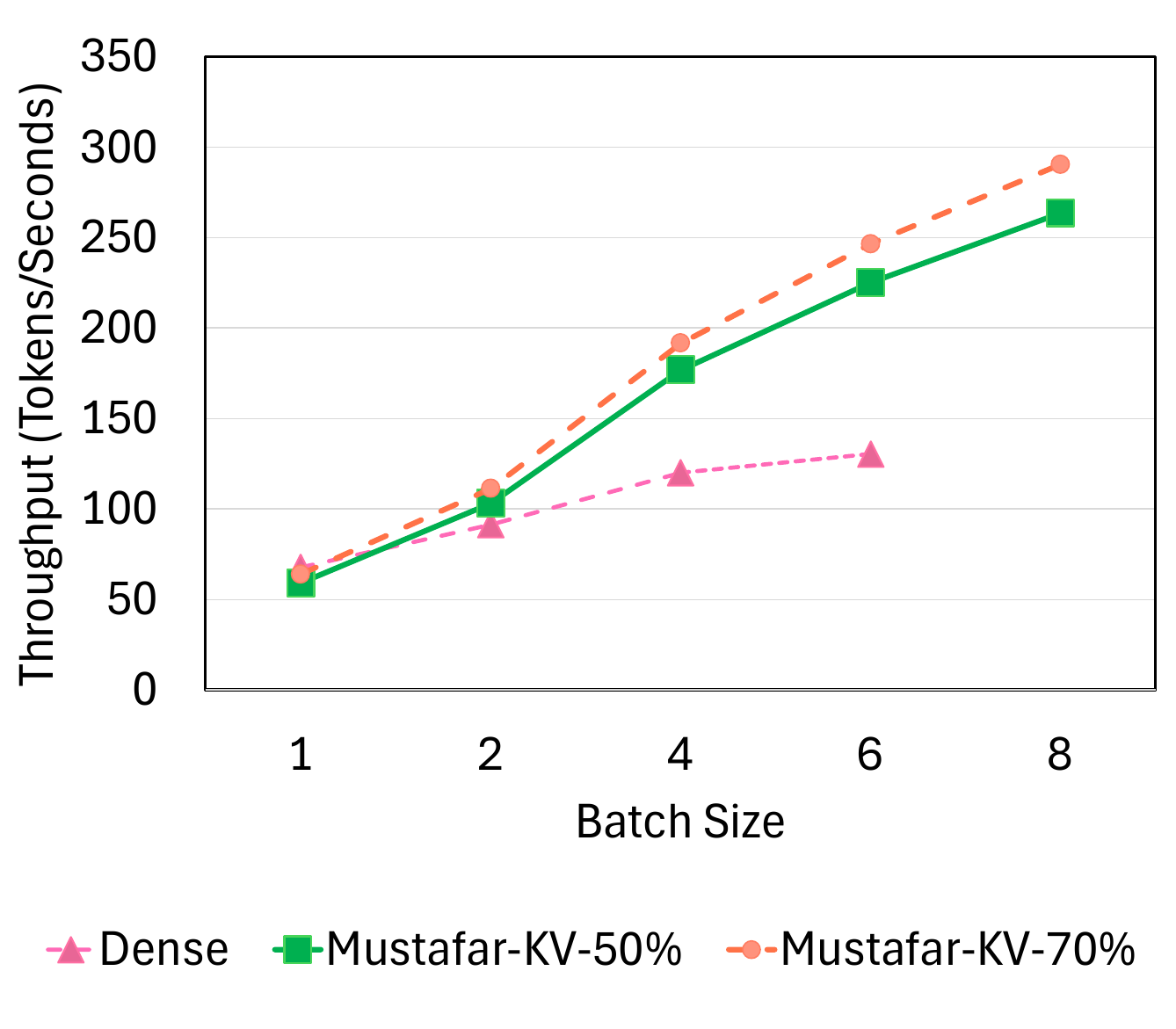}}
\end{subcaptionbox}
\caption{Throughput comparison of Mustafar to dense inference. } 
\label{fig:tput}
\end{figure}

Figure~\ref{fig:tput} shows the throughput comparison to inference with dense models. 
For Llama-2 7B, we used input sequence length of 2048 and generated 2048 tokens. For Llama-3 8B, we use input sequence length of 4096 and generated 4096 tokens. 
For dense baseline, FlashAttention~\cite{flashattention2} was used on prefill and decode phase.
Overall, we see that Mustafar is able to achieve higher throughput as well as support larger batch size owing to the reduced memory footprint of KV cache. In Llama-3, we see that enabling batch size of 8 leads to 2.23$\times$ tokens/sec throughput compared the dense inference of batch size 6.    
Even within the same batch size, we see an increased throughput upto 1.89$\times$. This is due to the pruning and compression overhead amortized by the speedup of Mustafar sparse attention kernel, leading to faster inference latency. However in batch size 1, we see that throughput is lower than dense inference. This is due to the underutilization of GPU in Mustafar sparse attention kernel with small batch size, where the number of threadblocks is smaller than the number of SMs. %We leave kernel optimization in this scienario as a future work.  
We provide additional throughput comparison with different input:output token ratios on Appendix~\ref{sec:appendix_kernel_eval}.

%Reducing the sequence length to run higher batch size, thereby increasing the size of KV cache, we see in detail the impact of KV cache compression. 
%Enables larger batch size, 

%\newpage
\section{Related Work}

\textbf{KV cache compression}
Alongside aforementioned work in KV cache pruning~\cite{think, lv2024kvpruner}, quantization~\cite{kivi, zipcache, 1bitkv, qhitter}, token-wise eviction~\cite{h2o, scissorhands, snapkv, less, keyformer, ge2024discard}, and low-rank approximation~\cite{asvd, palu, sun2025shadowkv, zhang2025tensorproduct, lin2024matryoshkakv}, 
KV cache offloading~\cite{infinigen, liu2024retrievalattention, hao2025omnikv, chen2024magicpiglshsamplingefficient} evicts KV cache to CPU memory and speculatively prefetchs critical tokens' KV cache. Layer-centric compression~\cite{zhang2025leankv, liu2024minicache} applies different level of compression to different layers, adhering to layer-wise importance. Head-level compression~\cite{fu2024headsmatter, tang2024razorattention} applies different level of compression to each heads, from the observation that not all heads contribute equally. Phase-specific compression~\cite{wu2024scope} applies different strategy for prefill and decode phase, with information retention prioritized in prefill and heavy hitter selection applied on decode phase. 

\textbf{System/Kernel for Attention}
While Mustafar attention kernel focuses on operating directly on the bitmap-compressed sparse KV cache, there exists various contributions from the system and kernel-levels to optimize for attention. 
PagedAttention~\cite{kwon2023pagedattention} introduces an paging-inspired attention algorithm that partitions KV cache into memory blocks to reduce memory fragmentation and efficient sharing across sequences.
FlashDecoding\cite{hong2024flashdecoding} introduces double-buffering to accelerate memory-bound GeMM of decode phase. 
FlashInfer~\cite{flashinfer} unifies KV cache format using a block-sparse representation for an efficient management of KV cache that leads to increased throughput.  
Loki~\cite{loki} uses a sparse attention method that leverages the low-dimensionality of key vectors to perform an approximate attention in a reduced PCA space.

\section{Conclusion and Limitations}

In this work, we demonstrate that unstructured sparsity presents a powerful and novel solution for KV cache pruning. By removing constraints on the pruning pattern, we show that per-token magnitude-based pruning achieves high sparsity while maintaining model accuracy.
To unlock the practical benefits of unstructured sparsity, we introduce a bitmap-based sparse format and a custom attention kernel that directly operates on compressed KV cache. 
Together, our pruning strategy, sparse format, and custom kernel form an end-to-end system that substantially reduces KV cache memory usage and improves throughput, making it possible to support longer contexts and more efficient inference. Mustafar establishes a foundation for future efforts to integrate unstructured sparsity into practical LLM deployment pipelines and opens new directions for memory-efficient LLM inference at scale. In future work, we plan to explore the joint effect of leveraging KV sparsity of Mustafar with sparsity in weights derived by works such as output-aware weight pruning~\cite{wanda}, pruning with low-rank adapters for accuracy retention~\cite{mozaffari2025slopedoubleprunedsparseplus, mozaffari2025slimoneshotquantizationsparsity}, and activation-aware calibration and efficiency enhancement~\cite{mirzadeh2023relustrikesbackexploiting, yin2025duogpttrainingfreedualsparsity}. Additionally, this paper focuses on showing that unstructure sparsity can prune both Key and Value caches to a higher sparsity with better accuracy than structured sparsity, leaving our method's ability to map arbitrary sparsity degree untouched. While we explore higher sparsity uniformly applied to the entire KV cache in Appendix~\ref{sec:appendix_higher_sparsity}, a future work involves deriving the optimal target sparsity to a smaller granularity (e.g. per-head or per-layer) to maximize sparsity and accuracy retention.

\begin{ack}
We gratefully acknowledge the support of National Science Foundation (NSF) under program PPoSS, Award Number 2316177.
\end{ack}

\newpage
%Phase 2
\bibliographystyle{plainnat}  % or another appropriate style

%phase1
%\bibliographystyle{plainnat}
%\bibliography{refs}

\begin{thebibliography}{53}
\providecommand{\natexlab}[1]{#1}
\providecommand{\url}[1]{\texttt{#1}}
\expandafter\ifx\csname urlstyle\endcsname\relax
  \providecommand{\doi}[1]{doi: #1}\else
  \providecommand{\doi}{doi: \begingroup \urlstyle{rm}\Url}\fi

\bibitem[Adnan et~al.(2024)Adnan, Arunkumar, Jain, Nair, Soloveychik, and Kamath]{keyformer}
Muhammad Adnan, Akhil Arunkumar, Gaurav Jain, Prashant~J. Nair, Ilya Soloveychik, and Purushotham Kamath.
\newblock Keyformer: Kv cache reduction through key tokens selection for efficient generative inference.
\newblock In P.~Gibbons, G.~Pekhimenko, and C.~De Sa, editors, \emph{Proceedings of Machine Learning and Systems}, volume~6, pages 114--127, 2024.
\newblock URL \url{https://proceedings.mlsys.org/paper_files/paper/2024/file/48fecef47b19fe501d27d338b6d52582-Paper-Conference.pdf}.

\bibitem[Ainslie et~al.(2023)Ainslie, Lee-Thorp, de~Jong, Zemlyanskiy, Lebron, and Sanghai]{ainslie2023gqa}
Joshua Ainslie, James Lee-Thorp, Michiel de~Jong, Yury Zemlyanskiy, Federico Lebron, and Sumit Sanghai.
\newblock {GQA}: Training generalized multi-query transformer models from multi-head checkpoints.
\newblock In \emph{The 2023 Conference on Empirical Methods in Natural Language Processing}, 2023.
\newblock URL \url{https://openreview.net/forum?id=hmOwOZWzYE}.

\bibitem[Bai et~al.(2024)Bai, Lv, Zhang, Lyu, Tang, Huang, Du, Liu, Zeng, Hou, Dong, Tang, and Li]{bai2024longbench}
Yushi Bai, Xin Lv, Jiajie Zhang, Hongchang Lyu, Jiankai Tang, Zhidian Huang, Zhengxiao Du, Xiao Liu, Aohan Zeng, Lei Hou, Yuxiao Dong, Jie Tang, and Juanzi Li.
\newblock {L}ong{B}ench: A bilingual, multitask benchmark for long context understanding.
\newblock In Lun-Wei Ku, Andre Martins, and Vivek Srikumar, editors, \emph{Proceedings of the 62nd Annual Meeting of the Association for Computational Linguistics (Volume 1: Long Papers)}, pages 3119--3137, Bangkok, Thailand, August 2024. Association for Computational Linguistics.
\newblock \doi{10.18653/v1/2024.acl-long.172}.
\newblock URL \url{https://aclanthology.org/2024.acl-long.172/}.

\bibitem[Chang et~al.(2025)Chang, Lin, Lin, Chen, Hu, Wang, Huang, Ceze, Abdelfattah, and Wu]{palu}
Chi-Chih Chang, Wei-Cheng Lin, Chien-Yu Lin, Chong-Yan Chen, Yu-Fang Hu, Pei-Shuo Wang, Ning-Chi Huang, Luis Ceze, Mohamed~S. Abdelfattah, and Kai-Chiang Wu.
\newblock Palu: {KV}-cache compression with low-rank projection.
\newblock In \emph{The Thirteenth International Conference on Learning Representations}, 2025.
\newblock URL \url{https://openreview.net/forum?id=LWMS4pk2vK}.

\bibitem[Chen et~al.(2025)Chen, Sadhukhan, Ye, Zhou, Zhang, Nolte, Tian, Douze, Bottou, Jia, and Chen]{chen2024magicpiglshsamplingefficient}
Zhuoming Chen, Ranajoy Sadhukhan, Zihao Ye, Yang Zhou, Jianyu Zhang, Niklas Nolte, Yuandong Tian, Matthijs Douze, Leon Bottou, Zhihao Jia, and Beidi Chen.
\newblock Magicpig: Lsh sampling for efficient llm generation.
\newblock In \emph{Proceedings of the 13th International Conference on Learning Representations (ICLR)}, 2025.
\newblock URL \url{https://arxiv.org/abs/2410.16179}.

\bibitem[Dao(2024)]{flashattention2}
Tri Dao.
\newblock Flashattention-2: Faster attention with better parallelism and work partitioning.
\newblock In \emph{Proceedings of the 12th International Conference on Learning Representations}, 2024.

\bibitem[DeepSeek-AI et~al.(2025)DeepSeek-AI, Guo, Yang, Zhang, Song, Zhang, Xu, et~al.]{deepseek2025r1}
DeepSeek-AI, Daya Guo, Dejian Yang, Haowei Zhang, Junxiao Song, Ruoyu Zhang, Runxin Xu, et~al.
\newblock Deepseek-r1: Incentivizing reasoning capability in llms via reinforcement learning, 2025.
\newblock URL \url{https://doi.org/10.48550/arXiv.2501.12948}.

\bibitem[Dong et~al.(2024)Dong, Yang, Zhang, Wang, Chi, and Chen]{less}
Harry Dong, Xinyu Yang, Zhenyu Zhang, Zhangyang Wang, Yuejie Chi, and Beidi Chen.
\newblock Get more with less: synthesizing recurrence with kv cache compression for efficient llm inference.
\newblock In \emph{Proceedings of the 41st International Conference on Machine Learning}, ICML'24. JMLR.org, 2024.

\bibitem[Fan et~al.(2025)Fan, Yu, Dong, Li, Gong, Wang, Wang, and Chu]{fan2025spinfer}
Ruibo Fan, Xiangrui Yu, Peijie Dong, Zeyu Li, Gu~Gong, Qiang Wang, Wei Wang, and Xiaowen Chu.
\newblock Spinfer: Leveraging low-level sparsity for efficient large language model inference on gpus.
\newblock In \emph{Proceedings of the Twentieth European Conference on Computer Systems (EuroSys '25)}, pages 243--260, New York, NY, USA, 2025. Association for Computing Machinery.
\newblock \doi{10.1145/3689031.3717481}.
\newblock URL \url{https://doi.org/10.1145/3689031.3717481}.

\bibitem[Fu et~al.(2025)Fu, Cai, Asi, Xiong, Dong, and Xiao]{fu2024headsmatter}
Yu~Fu, Zefan Cai, Abedelkadir Asi, Wayne Xiong, Yue Dong, and Wen Xiao.
\newblock Not all heads matter: A head-level {KV} cache compression method with integrated retrieval and reasoning.
\newblock In \emph{The Thirteenth International Conference on Learning Representations}, 2025.
\newblock URL \url{https://openreview.net/forum?id=FJFVmeXusW}.

\bibitem[Ge et~al.(2024)Ge, Zhang, Liu, Zhang, Han, and Gao]{ge2024discard}
Suyu Ge, Yunan Zhang, Liyuan Liu, Minjia Zhang, Jiawei Han, and Jianfeng Gao.
\newblock Model tells you what to discard: Adaptive kv cache compression for llms.
\newblock In \emph{Proceedings of the 2024 Conference on Empirical Methods in Natural Language Processing (EMNLP)}, 2024.

\bibitem[Grattafiori et~al.(2024)Grattafiori, Dubey, Jauhri, Pandey, Kadian, Al-Dahle, Letman, Mathur, Schelten, Vaughan, Yang, Fan, Goyal, Hartshorn, Yang, Mitra, Sravankumar, Korenev, Hinsvark, Rao, Zhang, Rodriguez, Gregerson, Spataru, Roziere, Biron, Tang, Chern, Caucheteux, Nayak, Bi, Marra, McConnell, Keller, Touret, Wu, Wong, Ferrer, Nikolaidis, Allonsius, Song, Pintz, Livshits, Wyatt, Esiobu, Choudhary, Mahajan, Garcia-Olano, Perino, Hupkes, Lakomkin, AlBadawy, Lobanova, Dinan, Smith, and many others]{grattafiori2024llama3}
Aaron Grattafiori, Abhimanyu Dubey, Abhinav Jauhri, Abhinav Pandey, Abhishek Kadian, Ahmad Al-Dahle, Aiesha Letman, Akhil Mathur, Alan Schelten, Alex Vaughan, Amy Yang, Angela Fan, Anirudh Goyal, Anthony Hartshorn, Aobo Yang, Archi Mitra, Archie Sravankumar, Artem Korenev, Arthur Hinsvark, Arun Rao, Aston Zhang, Aurelien Rodriguez, Austen Gregerson, Ava Spataru, Baptiste Roziere, Bethany Biron, Binh Tang, Bobbie Chern, Charlotte Caucheteux, Chaya Nayak, Chloe Bi, Chris Marra, Chris McConnell, Christian Keller, Christophe Touret, Chunyang Wu, Corinne Wong, Cristian~Canton Ferrer, Cyrus Nikolaidis, Damien Allonsius, Daniel Song, Danielle Pintz, Danny Livshits, Danny Wyatt, David Esiobu, Dhruv Choudhary, Dhruv Mahajan, Diego Garcia-Olano, Diego Perino, Dieuwke Hupkes, Egor Lakomkin, Ehab AlBadawy, Elina Lobanova, Emily Dinan, Eric~Michael Smith, and many others.
\newblock The llama 3 herd of models, 2024.
\newblock URL \url{https://arxiv.org/abs/2407.21783}.

\bibitem[Hao et~al.(2025)Hao, Zhu, Wang, Yu, Xin, Zheng, Ren, and Guo]{hao2025omnikv}
Jitai Hao, Yuke Zhu, Tian Wang, Jun Yu, Xin Xin, Bo~Zheng, Zhaochun Ren, and Sheng Guo.
\newblock Omnikv: Dynamic context selection for efficient long-context llms.
\newblock In \emph{Proceedings of the 13th International Conference on Learning Representations (ICLR)}, 2025.

\bibitem[Harma et~al.(2025)Harma, Chakraborty, Kostenok, Mishin, Ha, Falsafi, Jaggi, Liu, Oh, Subramanian, and Yazdanbakhsh]{harma2025interplay}
Simla~Burcu Harma, Ayan Chakraborty, Elizaveta Kostenok, Danila Mishin, Dongho Ha, Babak Falsafi, Martin Jaggi, Ming Liu, Yunho Oh, Suvinay Subramanian, and Amir Yazdanbakhsh.
\newblock Effective interplay between sparsity and quantization: From theory to practice.
\newblock In \emph{The Thirteenth International Conference on Learning Representations}, 2025.
\newblock URL \url{https://openreview.net/forum?id=wJv4AIt4sK}.

\bibitem[He et~al.(2024)He, Zhang, Wu, Liu, Zhou, and Zhuang]{zipcache}
Yefei He, Luoming Zhang, Weijia Wu, Jing Liu, Hong Zhou, and Bohan Zhuang.
\newblock Zipcache: Accurate and efficient {KV} cache quantization with salient token identification.
\newblock In \emph{The Thirty-eighth Annual Conference on Neural Information Processing Systems}, 2024.
\newblock URL \url{https://openreview.net/forum?id=5t4ZAkPiJs}.

\bibitem[Hong et~al.(2024)Hong, Dai, Xu, Mao, Li, Liu, Chen, Dong, and Wang]{hong2024flashdecoding}
Ke~Hong, Guohao Dai, Jiaming Xu, Qiuli Mao, Xiuhong Li, Jun Liu, Kangdi Chen, Yuhan Dong, and Yu~Wang.
\newblock Flashdecoding++: Faster large language model inference with asynchronization, flat gemm optimization, and heuristics.
\newblock In P.~Gibbons, G.~Pekhimenko, and C.~De Sa, editors, \emph{Proceedings of Machine Learning and Systems}, volume~6, pages 148--161, 2024.
\newblock URL \url{https://proceedings.mlsys.org/paper_files/paper/2024/file/5321b1dabcd2be188d796c21b733e8c7-Paper-Conference.pdf}.

\bibitem[Hsieh et~al.(2024)Hsieh, Sun, Kriman, Acharya, Rekesh, Jia, Zhang, and Ginsburg]{hsieh2024ruler}
Cheng-Ping Hsieh, Simeng Sun, Samuel Kriman, Shantanu Acharya, Dima Rekesh, Fei Jia, Yang Zhang, and Boris Ginsburg.
\newblock Ruler: What's the real context size of your long-context language models?
\newblock In \emph{Proceedings of COLM 2024}, 2024.

\bibitem[Huot et~al.(2025)Huot, Amplayo, Palomaki, Jakobovits, Clark, and Lapata]{huot2025agentsroom}
Fantine Huot, Reinald~Kim Amplayo, Jennimaria Palomaki, Alice~Shoshana Jakobovits, Elizabeth Clark, and Mirella Lapata.
\newblock Agents' room: Narrative generation through multi-step collaboration.
\newblock In \emph{The Thirteenth International Conference on Learning Representations}, 2025.
\newblock URL \url{https://openreview.net/forum?id=HfWcFs7XLR}.

\bibitem[Jiang et~al.(2023)Jiang, Sablayrolles, Mensch, Bamford, Chaplot, de~las Casas, Bressand, Lengyel, Lample, Saulnier, Lavaud, Lachaux, Stock, Scao, Lavril, Wang, Lacroix, and Sayed]{jiang2023mistral7b}
Albert~Q. Jiang, Alexandre Sablayrolles, Arthur Mensch, Chris Bamford, Devendra~Singh Chaplot, Diego de~las Casas, Florian Bressand, Gianna Lengyel, Guillaume Lample, Lucile Saulnier, Lélio~Renard Lavaud, Marie-Anne Lachaux, Pierre Stock, Teven~Le Scao, Thibaut Lavril, Thomas Wang, Timothée Lacroix, and William~El Sayed.
\newblock Mistral 7b.
\newblock In \emph{arXiv preprint arXiv:2310.06825}, 2023.
\newblock URL \url{https://arxiv.org/abs/2310.06825}.

\bibitem[Joo et~al.(2025)Joo, Hosseini, Hadidi, and Asgari]{joo2025coruscant}
Donghyeon Joo, Helya Hosseini, Ramyad Hadidi, and Bahar Asgari.
\newblock Coruscant: Co-designing gpu kernel and sparse tensor core to advocate unstructured sparsity in efficient llm inference.
\newblock In \emph{Proceedings of the 58th IEEE/ACM International Symposium on Microarchitecture}, 2025.
\newblock ISBN 9798400715730.
\newblock \doi{10.1145/3725843.3756065}.
\newblock URL \url{https://doi.org/10.1145/3725843.3756065}.

\bibitem[Jouppi et~al.(2023)Jouppi, Kurian, Li, Ma, Nagarajan, Nai, Patil, Subramanian, Swing, Towles, Young, Zhou, Zhou, and Patterson]{jouppi2023tpuv4}
Norman~P. Jouppi, George Kurian, Sheng Li, Peter Ma, Rahul Nagarajan, Lifeng Nai, Nishant Patil, Suvinay Subramanian, Andy Swing, Brian Towles, Cliff Young, Xiang Zhou, Zongwei Zhou, and David Patterson.
\newblock Tpu v4: An optically reconfigurable supercomputer for machine learning with hardware support for embeddings.
\newblock In \emph{Proceedings of the 50th Annual International Symposium on Computer Architecture (ISCA)}, 2023.
\newblock URL \url{https://arxiv.org/abs/2304.01433}.

\bibitem[Kim et~al.(2024)Kim, Chang, Karpinska, Garimella, Manjunatha, Lo, Goyal, and Iyyer]{kim2024fables}
Yekyung Kim, Yapei Chang, Marzena Karpinska, Aparna Garimella, Varun Manjunatha, Kyle Lo, Tanya Goyal, and Mohit Iyyer.
\newblock {FABLES}: Evaluating faithfulness and content selection in book-length summarization.
\newblock In \emph{First Conference on Language Modeling}, 2024.
\newblock URL \url{https://openreview.net/forum?id=YfHxQSoaWU}.

\bibitem[Kwon et~al.(2023)Kwon, Li, Zhuang, Sheng, Zheng, Yu, Gonzalez, Zhang, and Stoica]{kwon2023pagedattention}
Woosuk Kwon, Zhuohan Li, Siyuan Zhuang, Ying Sheng, Lianmin Zheng, Cody~Hao Yu, Joseph~E. Gonzalez, Hao Zhang, and Ion Stoica.
\newblock Efficient memory management for large language model serving with pagedattention.
\newblock In \emph{Proceedings of the 29th Symposium on Operating Systems Principles}, SOSP '23, page 611–626, New York, NY, USA, 2023. Association for Computing Machinery.
\newblock ISBN 9798400702297.
\newblock \doi{10.1145/3600006.3613165}.
\newblock URL \url{https://doi.org/10.1145/3600006.3613165}.

\bibitem[Lee et~al.(2024)Lee, Lee, Seo, and Sim]{infinigen}
Wonbeom Lee, Jungi Lee, Junghwan Seo, and Jaewoong Sim.
\newblock {InfiniGen}: Efficient generative inference of large language models with dynamic {KV} cache management.
\newblock In \emph{18th USENIX Symposium on Operating Systems Design and Implementation (OSDI 24)}, pages 155--172, Santa Clara, CA, July 2024. USENIX Association.
\newblock ISBN 978-1-939133-40-3.
\newblock URL \url{https://www.usenix.org/conference/osdi24/presentation/lee}.

\bibitem[Li et~al.(2024)Li, Huang, Yang, Venkitesh, Locatelli, Ye, Cai, Lewis, and Chen]{snapkv}
Yuhong Li, Yingbing Huang, Bowen Yang, Bharat Venkitesh, Acyr Locatelli, Hanchen Ye, Tianle Cai, Patrick Lewis, and Deming Chen.
\newblock Snap{KV}: {LLM} knows what you are looking for before generation.
\newblock In \emph{The Thirty-eighth Annual Conference on Neural Information Processing Systems}, 2024.
\newblock URL \url{https://openreview.net/forum?id=poE54GOq2l}.

\bibitem[Lin et~al.(2025)Lin, Zeng, Xiao, Kou, Hou, Gao, Zhang, and Deng]{lin2024matryoshkakv}
Bokai Lin, Zihao Zeng, Zipeng Xiao, Siqi Kou, TianQi Hou, Xiaofeng Gao, Hao Zhang, and Zhijie Deng.
\newblock Matryoshka{KV}: Adaptive {KV} compression via trainable orthogonal projection.
\newblock In \emph{The Thirteenth International Conference on Learning Representations}, 2025.
\newblock URL \url{https://openreview.net/forum?id=BQwsRy1h3U}.

\bibitem[Liu et~al.(2024{\natexlab{a}})Liu, Liu, Pan, He, Haffari, and Zhuang]{liu2024minicache}
Akide Liu, Jing Liu, Zizheng Pan, Yefei He, Gholamreza Haffari, and Bohan Zhuang.
\newblock Minicache: {KV} cache compression in depth dimension for large language models.
\newblock In \emph{The Thirty-eighth Annual Conference on Neural Information Processing Systems}, 2024{\natexlab{a}}.
\newblock URL \url{https://openreview.net/forum?id=sgVOjDqUMT}.

\bibitem[Liu et~al.(2024{\natexlab{b}})Liu, Chen, Lu, Jiang, Han, Zhang, Chen, Zhang, Ding, Zhang, Chen, Yang, Yang, and Qiu]{liu2024retrievalattention}
Di~Liu, Meng Chen, Baotong Lu, Huiqiang Jiang, Zhenhua Han, Qianxi Zhang, Qi~Chen, Chengruidong Zhang, Bailu Ding, Kai Zhang, Chen Chen, Fan Yang, Yuqing Yang, and Lili Qiu.
\newblock Retrievalattention: Accelerating long-context llm inference via vector retrieval, 2024{\natexlab{b}}.
\newblock URL \url{https://arxiv.org/abs/2409.10516}.

\bibitem[Liu et~al.(2023)Liu, Desai, Liao, Wang, Xie, Xu, Kyrillidis, and Shrivastava]{scissorhands}
Zichang Liu, Aditya Desai, Fangshuo Liao, Weitao Wang, Victor Xie, Zhaozhuo Xu, Anastasios Kyrillidis, and Anshumali Shrivastava.
\newblock Scissorhands: Exploiting the persistence of importance hypothesis for {LLM} {KV} cache compression at test time.
\newblock In \emph{Thirty-seventh Conference on Neural Information Processing Systems}, 2023.
\newblock URL \url{https://openreview.net/forum?id=JZfg6wGi6g}.

\bibitem[Liu et~al.(2024{\natexlab{c}})Liu, Yuan, Jin, Zhong, Xu, Braverman, Chen, and Hu]{kivi}
Zirui Liu, Jiayi Yuan, Hongye Jin, Shaochen Zhong, Zhaozhuo Xu, Vladimir Braverman, Beidi Chen, and Xia Hu.
\newblock {KIVI}: A tuning-free asymmetric 2bit quantization for {KV} cache.
\newblock In \emph{Forty-first International Conference on Machine Learning}, 2024{\natexlab{c}}.
\newblock URL \url{https://openreview.net/forum?id=L057s2Rq8O}.

\bibitem[Lv et~al.(2025)Lv, Zhou, Ding, Wang, and Ma]{lv2024kvpruner}
Bo~Lv, Quan Zhou, Xuanang Ding, Yan Wang, and Zeming Ma.
\newblock Kvpruner: Structural pruning for faster and memory-efficient large language models.
\newblock In \emph{ICASSP 2025 - 2025 IEEE International Conference on Acoustics, Speech and Signal Processing (ICASSP)}, pages 1--5, 2025.
\newblock \doi{10.1109/ICASSP49660.2025.10889000}.

\bibitem[Mirzadeh et~al.(2024)Mirzadeh, Alizadeh, Mehta, Mundo, Tuzel, Samei, Rastegari, and Farajtabar]{mirzadeh2023relustrikesbackexploiting}
Iman Mirzadeh, Keivan Alizadeh, Sachin Mehta, Carlo C~Del Mundo, Oncel Tuzel, Golnoosh Samei, Mohammad Rastegari, and Mehrdad Farajtabar.
\newblock Relu strikes back: Exploiting activation sparsity in large language models.
\newblock In \emph{Proceedings of the Twelfth International Conference on Learning Representations}, 2024.
\newblock URL \url{https://arxiv.org/pdf/2310.04564}.

\bibitem[Mozaffari et~al.(2025{\natexlab{a}})Mozaffari, Yazdanbakhsh, and Dehnavi]{mozaffari2025slimoneshotquantizationsparsity}
Mohammad Mozaffari, Amir Yazdanbakhsh, and Maryam~Mehri Dehnavi.
\newblock Slim: One-shot quantization and sparsity with low-rank approximation for llm weight compression.
\newblock In \emph{Proceedings of the 42nd International Conference on Machine Learning (ICML 2025)}, 2025{\natexlab{a}}.
\newblock URL \url{https://arxiv.org/abs/2410.09615}.

\bibitem[Mozaffari et~al.(2025{\natexlab{b}})Mozaffari, Yazdanbakhsh, Zhang, and Dehnavi]{mozaffari2025slopedoubleprunedsparseplus}
Mohammad Mozaffari, Amir Yazdanbakhsh, Zhao Zhang, and Maryam~Mehri Dehnavi.
\newblock Slope: Double-pruned sparse plus lazy low-rank adapter pretraining of llms.
\newblock In \emph{Proceedings of the International Conference on Learning Representations (ICLR 2025)}, 2025{\natexlab{b}}.
\newblock URL \url{https://arxiv.org/abs/2405.16325}.

\bibitem[Raihan et~al.(2019)Raihan, Goli, and Aamodt]{modelingstc}
Md~Aamir Raihan, Negar Goli, and Tor Aamodt.
\newblock Modeling deep learning accelerator enabled gpus, 2019.

\bibitem[Singhania et~al.(2024)Singhania, Singh, He, Feizi, and Bhatele]{loki}
Prajwal Singhania, Siddharth Singh, Shwai He, Soheil Feizi, and Abhinav Bhatele.
\newblock Loki: Low-rank keys for efficient sparse attention.
\newblock In \emph{The Thirty-eighth Annual Conference on Neural Information Processing Systems}, 2024.
\newblock URL \url{https://openreview.net/forum?id=raABeiV71j}.

\bibitem[Sun et~al.(2025)Sun, Chang, Bao, Zheng, Zheng, Liu, Dong, Chi, and Chen]{sun2025shadowkv}
Hanshi Sun, Li-Wen Chang, Wenlei Bao, Size Zheng, Ningxin Zheng, Xin Liu, Harry Dong, Yuejie Chi, and Beidi Chen.
\newblock Shadowkv: Kv cache in shadows for high-throughput long-context llm inference.
\newblock In \emph{Proceedings of the Forty-Second International Conference on Machine Learning}, 2025.
\newblock URL \url{https://arxiv.org/abs/2410.21465}.

\bibitem[Sun et~al.(2024)Sun, Liu, Bair, and Kolter]{wanda}
Mingjie Sun, Zhuang Liu, Anna Bair, and J.~Zico Kolter.
\newblock A simple and effective pruning approach for large language models.
\newblock In \emph{Proceedings of the 12th International Conference on Learning Representations}, 2024.

\bibitem[Tang et~al.(2024)Tang, Lin, Lin, Han, Hong, Yao, and Wang]{tang2024razorattention}
Hanlin Tang, Yang Lin, Jing Lin, Qingsen Han, Shikuan Hong, Yiwu Yao, and Gongyi Wang.
\newblock Razorattention: Efficient kv cache compression through retrieval heads.
\newblock In \emph{arXiv preprint arXiv:2407.15891}, 2024.
\newblock URL \url{https://arxiv.org/abs/2407.15891}.

\bibitem[Touvron et~al.(2023)Touvron, Martin, Stone, Albert, Almahairi, Babaei, Bashlykov, Batra, Bhargava, Bhosale, Bikel, Blecher, Ferrer, Chen, Cucurull, Esiobu, Fernandes, Fu, Fu, Fuller, Gao, Goswami, Goyal, Hartshorn, Hosseini, Hou, Inan, Kardas, Kerkez, Khabsa, Kloumann, Korenev, Koura, Lachaux, Lavril, Lee, Liskovich, Lu, Mao, Martinet, Mihaylov, Mishra, Molybog, Nie, Poulton, Reizenstein, Rungta, Saladi, Schelten, Silva, Smith, Subramanian, Tan, Tang, Taylor, Williams, Kuan, Xu, Yan, Zarov, Zhang, Fan, Kambadur, Narang, Rodriguez, Stojnic, Edunov, and Scialom]{touvron2023llama2}
Hugo Touvron, Louis Martin, Kevin Stone, Peter Albert, Amjad Almahairi, Yasmine Babaei, Nikolay Bashlykov, Soumya Batra, Prajjwal Bhargava, Shruti Bhosale, Dan Bikel, Lukas Blecher, Cristian~Canton Ferrer, Moya Chen, Guillem Cucurull, David Esiobu, Jude Fernandes, Jeremy Fu, Wenyin Fu, Brian Fuller, Cynthia Gao, Vedanuj Goswami, Naman Goyal, Anthony Hartshorn, Saghar Hosseini, Rui Hou, Hakan Inan, Marcin Kardas, Viktor Kerkez, Madian Khabsa, Isabel Kloumann, Artem Korenev, Punit~Singh Koura, Marie-Anne Lachaux, Thibaut Lavril, Jenya Lee, Diana Liskovich, Yinghai Lu, Yuning Mao, Xavier Martinet, Todor Mihaylov, Pushkar Mishra, Igor Molybog, Yixin Nie, Andrew Poulton, Jeremy Reizenstein, Rashi Rungta, Kalyan Saladi, Alan Schelten, Ruan Silva, Eric~Michael Smith, Ranjan Subramanian, Xiaoqing~Ellen Tan, Binh Tang, Ross Taylor, Adina Williams, Jian~Xiang Kuan, Puxin Xu, Zheng Yan, Iliyan Zarov, Yuchen Zhang, Angela Fan, Melanie Kambadur, Sharan Narang, Aurelien Rodriguez, Robert Stojnic, Sergey Edunov, and Thomas
  Scialom.
\newblock Llama 2: Open foundation and fine-tuned chat models, 2023.
\newblock URL \url{https://arxiv.org/abs/2307.09288}.

\bibitem[Vaswani et~al.(2017)Vaswani, Shazeer, Parmar, Uszkoreit, Jones, Gomez, Kaiser, and Polosukhin]{vaswani2017attention}
Ashish Vaswani, Noam Shazeer, Niki Parmar, Jakob Uszkoreit, Llion Jones, Aidan~N. Gomez, Lukasz Kaiser, and Illia Polosukhin.
\newblock Attention is all you need.
\newblock In \emph{Proceedings of the 31st International Conference on Neural Information Processing Systems (NeurIPS)}, 2017.
\newblock URL \url{https://doi.org/10.48550/arXiv.1706.03762}.

\bibitem[Wu et~al.(2025)Wu, Wang, Zhang, Lai, He, and Zhou]{wu2024scope}
Jialong Wu, Zhenglin Wang, Linhai Zhang, Yilong Lai, Yulan He, and Deyu Zhou.
\newblock Scope: Optimizing key-value cache compression in long-context generation.
\newblock In \emph{Proceedings of the 63rd Annual Meeting of the Association for Computational Linguistics}, 2025.
\newblock URL \url{https://aclanthology.org/2025.acl-long.529.pdf}.

\bibitem[Xia et~al.()Xia, Zheng, Li, Zhuang, Zhou, Qiu, Li, Lin, and Song]{flashllm}
Haojun Xia, Zhen Zheng, Yuchao Li, Donglin Zhuang, Zhongzhu Zhou, Xiafei Qiu, Yong Li, Wei Lin, and Shuaiwen~Leon Song.
\newblock Flash-llm: Enabling cost-effective and highly-efficient large generative model inference with unstructured sparsity.
\newblock volume~17. VLDB Endowment.
\newblock \doi{10.14778/3626292.3626303}.
\newblock URL \url{https://doi.org/10.14778/3626292.3626303}.

\bibitem[Xu et~al.(2025)Xu, Jie, Dong, Wang, Lu, Zhou, Saha, Xiong, and Sahoo]{think}
Yuhui Xu, Zhanming Jie, Hanze Dong, Lei Wang, Xudong Lu, Aojun Zhou, Amrita Saha, Caiming Xiong, and Doyen Sahoo.
\newblock Think: Thinner key cache by query-driven pruning.
\newblock In \emph{The Thirteenth International Conference on Learning Representations}, 2025.
\newblock URL \url{https://openreview.net/forum?id=n0OtGl6VGb}.

\bibitem[Ye et~al.(2025)Ye, Chen, Lai, Lin, Zhang, Wang, Chen, et~al.]{flashinfer}
Zihao Ye, Lequn Chen, Ruihang Lai, Wuwei Lin, Yineng Zhang, Stephanie Wang, Tianqi Chen, et~al.
\newblock Flashinfer: Efficient and customizable attention engine for llm inference serving, 2025.

\bibitem[Yin et~al.(2025)Yin, Li, Lee, and Panda]{yin2025duogpttrainingfreedualsparsity}
Ruokai Yin, Yuhang Li, Donghyun Lee, and Priyadarshini Panda.
\newblock Duogpt: Training-free dual sparsity through activation-aware pruning in llms, 2025.
\newblock URL \url{https://arxiv.org/abs/2506.20194}.

\bibitem[Yuan et~al.()Yuan, Shang, Song, Wu, Yan, and Sun]{asvd}
Zhihang Yuan, Yuzhang Shang, Yue Song, Qiang Wu, Yan Yan, and Guangyu Sun.
\newblock Asvd: Activation-aware singular value decomposition for compressing large language models.
\newblock URL \url{https://doi.org/10.48550/arXiv.2312.05821}.
\newblock arXiv preprint arXiv:2312.05821.

\bibitem[Zhang et~al.(2024{\natexlab{a}})Zhang, Yi, Xu, and Shrivastava]{1bitkv}
Tianyi Zhang, Jonah~Wonkyu Yi, Zhaozhuo Xu, and Anshumali Shrivastava.
\newblock {KV} cache is 1 bit per channel: Efficient large language model inference with coupled quantization.
\newblock In \emph{The Thirty-eighth Annual Conference on Neural Information Processing Systems}, 2024{\natexlab{a}}.
\newblock URL \url{https://openreview.net/forum?id=pNnvzQsS4P}.

\bibitem[Zhang et~al.(2024{\natexlab{b}})Zhang, Hu, Zhao, Lui, and Chen]{zhang2025leankv}
Yanqi Zhang, Yuwei Hu, Runyuan Zhao, John C.~S. Lui, and Haibo Chen.
\newblock Unifying kv cache compression for large language models with leankv, 2024{\natexlab{b}}.
\newblock URL \url{https://doi.org/10.48550/arXiv.2412.03131}.

\bibitem[Zhang et~al.(2025)Zhang, Liu, Yuan, Qin, Yuan, Gu, and Yao]{zhang2025tensorproduct}
Yifan Zhang, Yifeng Liu, Huizhuo Yuan, Zhen Qin, Yang Yuan, Quanquan Gu, and Andrew Chi-Chih Yao.
\newblock Tensor product attention is all you need, 2025.
\newblock URL \url{https://arxiv.org/abs/2501.06425}.

\bibitem[Zhang et~al.(2023)Zhang, Sheng, Zhou, Chen, Zheng, Cai, Song, Tian, Re, Barrett, Wang, and Chen]{h2o}
Zhenyu Zhang, Ying Sheng, Tianyi Zhou, Tianlong Chen, Lianmin Zheng, Ruisi Cai, Zhao Song, Yuandong Tian, Christopher Re, Clark Barrett, Zhangyang Wang, and Beidi Chen.
\newblock H2o: Heavy-hitter oracle for efficient generative inference of large language models.
\newblock In \emph{Thirty-seventh Conference on Neural Information Processing Systems}, 2023.
\newblock URL \url{https://openreview.net/forum?id=RkRrPp7GKO}.

\bibitem[Zhang et~al.(2024{\natexlab{c}})Zhang, Liu, Chen, Kailkhura, Chen, and Wang]{qhitter}
Zhenyu Zhang, Shiwei Liu, Runjin Chen, Bhavya Kailkhura, Beidi Chen, and Atlas Wang.
\newblock Q-hitter: A better token oracle for efficient llm inference via sparse-quantized kv cache.
\newblock In \emph{MLSys}, 2024{\natexlab{c}}.
\newblock URL \url{https://proceedings.mlsys.org/paper_files/paper/2024/hash/bbb7506579431a85861a05fff048d3e1-Abstract-Conference.html}.

\bibitem[Zhao et~al.(2025)Zhao, Jiang, Lee, Chiu, Cardie, Gall{\'e}, and Rush]{zhao2024commit0}
Wenting Zhao, Nan Jiang, Celine Lee, Justin~T Chiu, Claire Cardie, Matthias Gall{\'e}, and Alexander~M Rush.
\newblock Commit0: Library generation from scratch.
\newblock In \emph{The Thirteenth International Conference on Learning Representations}, 2025.
\newblock URL \url{https://openreview.net/forum?id=MMwaQEVsAg}.

\end{thebibliography}

\newpage 
\appendix
\section{Extended Evaluation}
\subsection{Section~\ref{sec:2} Methodology Applied to LLaMA-2 7B}
\label{appendix:llama2}
We follow the same methodology of exploring pruning direction and output-awareness on Llama-2-7B to further solidify our findings on a model architecture with Multi-Head Attention. 
In Table~\ref{tab:k_compare_appendix}, we observe a similar trend to that of Llama-3-8B-Instruct in Section~\ref{sec:2}. Unstructured pruning outperforms structured pruning of ThinK~\cite{think}, with ouput-awareness bringing a small accuracy increase to pure magnitude-based pruning. 

\begin{table}[h!]
\centering
\caption{Comparison of ThinK~\cite{think} structured pruning, per-token magnitude-based unstructured pruning, and per-token output-aware unstructured pruning on LongBench~\cite{bai2024longbench} with Llama-2-7B Key cache.}
\label{tab:k_compare_appendix}
\vspace{0.5em}
\resizebox{\textwidth}{!}{%
\begin{tabular}{|c|c|ccc|ccc|}
\hline
%\multirow{2}{*}{Dataset} & \textbf{Dense} & \multicolumn{2}{c|}{K = 0.5} & \multicolumn{2}{c|}{K = 0.7} \\
\multirow{3}{*}{\centering\textbf{Task}} & \multirow{3}{*}{\centering\textbf{Dense}} 
& \multicolumn{3}{c|}{$K_s$ = 0.5} & \multicolumn{3}{c|}{$K_s$ = 0.7}  \\
\cline{3-8}
%& & ThinK (Structured) & Mustafar (Unstructured) & ThinK (Structured) & Mustafar (Unstructured) \\
& & \makecell{ThinK \\ (Structured)} & \makecell{Unstructured \\ Output-aware} & \makecell{Unstructured \\ Magnitude} & \makecell{ThinK \\ (Structured)} & \makecell{Unstructured \\ Output-aware} & \makecell{Unstructured \\ Magnitude} \\
\hline
Average       & 27.51   & 25.70 & \textbf{27.55} & 27.46     & 21.57  & \textbf{26.78} & 26.17 \\
\hline
SingleDoc QA  & 15.53   & 16.28 & \textbf{15.52} & 15.37     & 14.17  & \textbf{15.82} & 14.43 \\
MultiDoc QA   & 7.10   & 6.30 & 6.90 & \textbf{7.23}     & 4.33  & 6.44 & \textbf{6.62} \\
Summarization & 14.02   & 7.46 & \textbf{14.51} & 13.91     & 9.28  & \textbf{12.99} & 11.64 \\
Few-shot      & 65.13   & 64.34 & \textbf{65.20} & 65.00     & 57.81  & \textbf{63.77} & 63.65 \\
Synthetic     & 4.17   & 2.42 & \textbf{3.98} & 3.90     & 4.35  & 3.00 & \textbf{3.08} \\
Code          & 63.24   & 61.57 & 63.22 & \textbf{63.52}     & 39.85  & \textbf{62.67} & 61.73 \\
\hline
\end{tabular}
}
\end{table}

In Table~\ref{tab:v_compare_appendix}, a unique phenomenon is the stark contrast of model accuracy in per-channel unstructured pruning methods. Whereas per-channel magnitude-based pruning of Table~\ref{tab:v_compare} show good model accuracy retention for Llama-3-8B-Instruct, for Llama-2-7B we see that accuracy degradation is very severe. Nevertheless, concurrent to our previous finding, we once again see that per-channel pruning achieves the same level of accuracy retention to per-token pruning as output-awareness is applied. This highlights the importance of output-awareness in Value cache pruning. In Table~\ref{tab:kv_compare_appendix} we see that the model accuracy of 70\% unstructured sparsity on both Key and Value cache achieves similar accuracy to 50\% ThinK pruning.  

\begin{table}[h!]
\centering
\caption{Comparison of ThinK~\cite{think} structured pruning, per-channel magnitude-based unstructured pruning, per-channel output-aware unstructured pruning, and per-token magnitude-based pruning on LongBench~\cite{bai2024longbench} with Llama-2-7B Value cache.}
\label{tab:v_compare_appendix}
\vspace{0.5em}
\resizebox{\textwidth}{!}{%
\begin{tabular}{|c|c|cccc|cccc|}
\hline
\multirow{3}{*}{\centering\textbf{Task}} & \multirow{3}{*}{\centering\textbf{Dense}} 
& \multicolumn{4}{c|}{$V_s$ = 0.5} & \multicolumn{4}{c|}{$V_s$ = 0.7}  \\
\cline{3-10}
& & \makecell{ThinK \\ (Structured)} & \makecell{Magnitude \\ (Per-channel)} & \makecell{Output-aware \\ (Per-channel)} & \makecell{Magnitude \\ (Per-token)}  & \makecell{ThinK \\ (Structured)} & \makecell{Magnitude \\ (Per-channel)} & \makecell{Output-aware \\ (Per-channel)} & \makecell{Magnitude \\ (Per-token)} \\
\hline
Average       & 27.51  & 24.59 & 6.16 & 27.33 & \textbf{27.39}     & 21.10 & 5.81 & 26.30 & \textbf{27.05} \\
\hline
SingleDoc QA  & 15.53  & 12.64 & 1.68 & \textbf{15.96} & 15.62     & 10.05 & 1.60 & \textbf{15.48} & 15.17 \\
MultiDoc QA   & 7.10  & 7.37 & 2.17 & \textbf{6.97} & 6.92     & 7.15 & 1.82 & 6.97 & \textbf{7.10} \\
Summarization & 14.02  & 9.18 & 4.51 & \textbf{13.98} & 13.54     & 9.10 & 3.15 & \textbf{13.06} & 12.67 \\
Few-shot      & 65.13  & 61.82 & 9.93 & 64.07 & \textbf{64.92}     & 57.12 & 8.83 & 60.09 & \textbf{64.80} \\
Synthetic     & 4.17  & 3.86 & 1.82 & \textbf{4.45} & 4.20     & 1.65 & 2.45 & \textbf{4.69} & 3.95 \\
Code          & 63.24  & 56.31 & 20.03 & 62.72 & \textbf{63.44}     & 41.96 & 20.90 & 62.34 & \textbf{62.85} \\
\hline
\end{tabular}
}
\end{table}

\begin{table}[h!]
\centering
\caption{Longbench evaluation of Llama-2 7B with KV cache per-token magnitude-based pruning}
\label{tab:kv_compare_appendix}
\vspace{0.5em}
\resizebox{0.40\linewidth}{!}{%
\begin{tabular}{|c|ccc|}
\hline
\multirow{3}{*}{Task} & \multicolumn{3}{c|}{Llama-2-7B}  \\
\cline{2-4}
&  \multirow{1}{*}{\textbf{Dense}} & \makecell{$K_s$ = 0.5 \\ $V_s$ = 0.5} & \makecell{$K_s$ = 0.7 \\ $V_s$ = 0.7}  \\
\hline
Average        & 27.51 & 27.23 & 24.71   \\
\hline
SingleDoc QA   & 15.53 & 15.21 & 13.62   \\
MultiDoc QA    & 7.10 & 7.11 & 6.78   \\
Summarization  & 14.02 & 13.61 & 6.84   \\
Few-shot       & 65.13 & 64.84 & 62.59   \\
Synthetic      & 4.17 & 3.55 & 2.63   \\
Code           & 63.24 & 63.16 & 60.35   \\
\hline
\end{tabular}
}
\end{table}

\newpage
\subsection{Scaling to Larger Model}
In Table~\ref{tab:llama2-13b}, we include the accuracy evaluation of Mustafar per-token magnitude-based pruning on Llama-2-13B-chat~\cite{touvron2023llama2}, validating the effectiveness of Mustafar on model with larger size. While unstructured pruning constantly outperforms structured sparsity, we see that the Key cache of Llama-2-13B-chat is more susceptible to accuracy degradation at 70\% sparsity (yellow). In this case, we leverage the modularity of Mustafar, being able to apply different target sparsity to Key and Value cache to find the best combination, to use 50\% sparsity for Key Cache and 70\% sparsity for Value cache (pink), thereby reaching the higher overall sparsity while maintaining the model accuracy.

\begin{table}[ht]
\centering
\caption{Mustafar accuracy with Llama-2-13B-chat on LongBench}
\label{tab:llama2-13b}
\vspace{0.5em}
\resizebox{\textwidth}{!}{%
\begin{tabular}{c|ccc|ccc|ccc|ccc|cc|cc|c}
\toprule
 & \multicolumn{3}{c|}{\textbf{Single-Document QA}} & \multicolumn{3}{c|}{\textbf{Multi-Document QA}} & \multicolumn{3}{c|}{\textbf{Summarization}} & \multicolumn{3}{c|}{\textbf{Few-shot Learning}} & \multicolumn{2}{c|}{\textbf{Synthetic}} & \multicolumn{2}{c|}{\textbf{Code}} &  \\
\cline{2-17}  
%\multirow[-0.5ex]{2}{*}\makecell{\textbf{KV} \\ \textbf{Sparsity}}
%\multirowcell{2}[0ex][c]{\textbf{KV} \\ \textbf{Sparsity} }
\makecell[c]{\textbf{KV}\\\textbf{Sparsity}}
%\makecell{\textbf{KV} \\ \textbf{Sparsity}}
& \makecell{\rotatebox{60}{\textbf{NtrvQA}}}
& \makecell{\rotatebox{60}{\textbf{Qasper}}}
& \makecell{\rotatebox{60}{\textbf{MF-en}}}
& \makecell{\rotatebox{60}{\textbf{HotpotQA}}}
& \makecell{\rotatebox{60}{\textbf{2WikiMQA}}}
& \makecell{\rotatebox{60}{\textbf{Musique}}}
& \makecell{\rotatebox{60}{\textbf{GovReport}}}
& \makecell{\rotatebox{60}{\textbf{QMSum}}}
& \makecell{\rotatebox{60}{\textbf{MultiNews}}}
& \makecell{\rotatebox{60}{\textbf{TREC}}}
& \makecell{\rotatebox{60}{\textbf{TrivialQA}}}
& \makecell{\rotatebox{60}{\textbf{SAMSum}}}
& \makecell{\rotatebox{60}{\textbf{PCount}}}
& \makecell{\rotatebox{60}{\textbf{PRe}}}
& \makecell{\rotatebox{60}{\textbf{Lec}}}
& \makecell{\rotatebox{60}{\textbf{RBP}}}
% & \textbf{Avg.}  \\
& \makecell[c]{\textbf{Avg.}} \\
\midrule
\multicolumn{18}{c}{\textbf{Llama-2-13B-Chat}} \\
Dense & 18.54 & 24.09 & 37.01 & 36.43 & 31.40 & 15.81 & 24.48 & 20.25 & 25.74 & 67.50 & 86.90 & 42.07 & 3.00 & 12.00 & 50.12 & 50.53 & \textbf{34.12} \\
ThinK0.5  & 16.95 & 22.39 & 37.54 & 34.00 & 29.93 & 14.33 & 24.49 & 20.21 & 24.78 & 67.50  & 87.16 & 40.53 & 2.55 & 13.07 & 45.79 & 46.23 & \textbf{32.80} \\
K0.5 V0.0   & 18.46 & 23.12 & 37.26 & 37.16 & 31.18 & 15.56 & 23.90 & 20.55 & 25.57 & 67.50 & 87.23 & 41.99 & 3.00 & 11.50 & 50.33 & 48.88 & \textbf{33.95} \\
ThinK0.7  & 17.86 & 19.93 & 32.37 & 33.03 & 27.22 & 13.99 & 21.19 & 19.47 & 12.04 & 59.0  & 86.67 & 31.26 & 1.54 & 1.87  & 27.79 & 29.35 & \textbf{27.16} \\
K0.7 V0.0  & 14.63 & 20.97 & 34.05 & 34.70 & 30.69 & 13.72 & 10.60 & 20.01 & 7.63 & 61.00 & 81.91 & 37.76 & 1.00 & 1.00 & 45.29 & 33.54 & \cellcolor{yellow}\textbf{28.03} \\
K0.0 V0.5 & 18.75 & 23.68 & 37.34 & 36.83 & 31.36 & 15.50  & 23.97 & 20.83 & 25.46 & 67.50  & 87.20  & 41.45  & 2.50  & 10.00    & 49.32 & 49.37 & \textbf{33.82} \\
K0.0 V0.7  & 19.29 & 22.90  & 37.65 & 36.57 & 31.24 & 15.35 & 22.44 & 20.52 & 24.75 & 68.00  & 87.49 & 40.55 & 2.50  & 8.10   & 49.33 & 49.14 & \textbf{33.49} \\
K0.5 V0.5 & 19.08 & 22.66 & 36.97 & 37.25 & 31.38 & 15.46  & 23.70 & 20.66 & 25.39 & 67.50  & 87.23 & 40.59  & 3.00  & 10.10  & 49.39 & 48.06 & \textbf{33.64} \\
K0.5 V0.7 & 18.60 & 22.57 & 37.18 & 35.40 & 31.55 & 15.25 & 22.30 & 20.43 & 24.81 & 68.00  & 87.23 & 39.91 & 2.50  & 7.70  & 49.02 & 47.38 & \cellcolor{pink}\textbf{33.24} \\
K0.7 V0.7  & 17.86 & 19.93 & 32.37 & 33.03 & 27.22 & 13.99 & 21.19 & 19.47 & 12.04 & 59.00  & 86.67 & 31.26 & 1.54 & 1.87  & 27.79 & 29.35 & \cellcolor{yellow}\textbf{27.16} \\
\bottomrule
\end{tabular}%
}
\end{table}

\subsection{Evaluation on RULER}
\label{sec:appendix_ruler}
For a more diverse evaluation, we evaluate Llama-3.1-8B-Instruct on RULER~\cite{hsieh2024ruler} benchmark for context length of 65,536 tokens. 

\begin{table}[h]
\centering
\caption{Accuracy comparison on RULER benchmark}
\label{tab:needles_qa}
\vspace{0.5em}
\resizebox{\textwidth}{!}{%
\begin{tabular}{c|c|ccccccccccc}
\toprule
\makecell[c]{\textbf{Sparsity}} &
\makecell[c]{\textbf{Method}} &
\makecell{\rotatebox{60}{\textbf{Needle-Single1}}} &
\makecell{\rotatebox{60}{\textbf{Needle-Single2}}} &
\makecell{\rotatebox{60}{\textbf{Needle-MultiKey1}}} &
\makecell{\rotatebox{60}{\textbf{Needle-MultiKey2}}} &
\makecell{\rotatebox{60}{\textbf{Needle-MultiQuery}}} &
\makecell{\rotatebox{60}{\textbf{Needle-MultiValue}}} &
\makecell{\rotatebox{60}{\textbf{QA-1}}} &
\makecell{\rotatebox{60}{\textbf{QA-2}}} &
\makecell{\rotatebox{60}{\textbf{Variable Tracking}}} &
\makecell{\rotatebox{60}{\textbf{Freq. Words Extract.}}} \\
\midrule
\multicolumn{12}{c}{\textbf{Llama-3.1-8B-Instruct}} \\
Dense & --- & 1.000 & 1.000 & 0.990 & 0.979 & 0.990 & 0.979 & 0.844 & 0.594 & 0.973 & 0.851 \\
\midrule

Key 50\% & ThinK & 1.000 & 1.000 & 0.990 & 0.979 & 0.995 & 0.969 & 0.833 & 0.594 & 0.919 & 0.854 \\
         & Mustafar & 1.000 & 1.000 & 0.990 & 0.979 & 0.995 & 0.996 & 0.833 & 0.573 & 0.971 & 0.813 \\
\midrule

Key 70\% & ThinK & 0.448 & 0.490 & 0.229 & 0.188 & 0.646 & 0.487 & 0.615 & 0.510 & 0.208 & 0.427 \\
         & Mustafar & 1.000 & 1.000 & 0.990 & 0.969 & 0.992 & 0.903 & 0.833 & 0.594 & 0.966 & 0.823 \\
\midrule

Value 50\% & ThinK & 1.000 & 1.000 & 0.990 & 0.969 & 0.914 & 0.958 & 0.823 & 0.573 & 0.910 & 0.792 \\
           & Mustafar & 1.000 & 1.000 & 0.979 & 0.995 & 0.995 & 0.971 & 0.833 & 0.604 & 0.983 & 0.830 \\
\midrule

Value 70\% & ThinK & 0.948 & 0.927 & 0.948 & 0.510 & 0.698 & 0.688 & 0.646 & 0.500 & 0.558 & 0.677 \\
           & Mustafar & 1.000 & 1.000 & 1.000 & 0.979 & 0.992 & 0.969 & 0.833 & 0.594 & 0.985 & 0.826 \\
\midrule

Key\&Value 50\% & ThinK & 0.958 & 1.000 & 0.948 & 0.854 & 0.828 & 0.956 & 0.740 & 0.531 & 0.742 & 0.823 \\
                & Mustafar & 1.000 & 1.000 & 0.990 & 0.979 & 0.997 & 0.997 & 0.833 & 0.573 & 0.862 & 0.809 \\
\midrule

Key\&Value 70\% & ThinK & 0.000 & 0.073 & 0.000 & 0.000 & 0.000 & 0.000 & 0.219 & 0.250 & 0.000 & 0.035 \\
                & Mustafar & 1.000 & 1.000 & 0.990 & 0.969 & 0.995 & 0.914 & 0.833 & 0.583 & 0.869 & 0.799 \\
\bottomrule
\end{tabular}%
}
\end{table}

As shown in Table~\ref{tab:needles_qa}, even in the challenging Needle-in-a-Haystack scenarios with multiple keys and queries, Mustafar maintains accuracy comparable to the dense model. It also outperforms the structured pruning baseline ThinK, with particularly notable gains at 70\% joint Key-Value sparsity. While structured pruning does perform well in isolated cases, such as the Needle-Single tasks for 70\% Value sparsity, it exhibits significant accuracy drops in other tasks. In contrast, Mustafar’s unstructured sparsity consistently preserves accuracy across all tasks. This contrast highlights the versatility of unstructured sparsity in adapting to diverse task requirements.

\newpage
\subsection{Higher Sparsity}
\label{sec:appendix_higher_sparsity}
While the main paper primarily focused on 50\% and 70\% sparsity of both Key and Value Cache, we present the performance of Mustafar per-token magnitude-based pruning of KV cache 80\% and 90\% sparsity in Table~\ref{tab:higher_sparsity}. While we see that Key cache suffers from accuracy degradation in higher sparsity, Value cache, despite the even distribution of element magnitude as in Figure~\ref{fig:sub2}, retains some level of the model accuracy even at 90\% sparsity on selective tasks. Model accuracy is retained for tasks such as 2WikiMultihopQA (pink), while degraded significantly in tasks such as GovReport (yellow).  

\begin{table}[ht]
\centering
\caption{Mustafar accuracy with Llama-3-8B-Instruct on LongBench}
\label{tab:higher_sparsity}
\vspace{0.5em}
\resizebox{\textwidth}{!}{%
\begin{tabular}{c|ccc|ccc|ccc|ccc|cc|cc|c}
\toprule
%\makecell{\textbf{KV} \\ \textbf{Sparsity}} & \multicolumn{3}{c|}{\textbf{Single-Document QA}} & \multicolumn{3}{c|}{\textbf{Multi-Document QA}} & \multicolumn{3}{c|}{\textbf{Summarization}} & \multicolumn{3}{c|}{\textbf{Few-shot Learning}} & \multicolumn{2}{c|}{\textbf{Synthetic}} & \multicolumn{2}{c|}{\textbf{Code}} & \textbf{Avg.} \\
%\midrule  
 & \multicolumn{3}{c|}{\textbf{Single-Document QA}} & \multicolumn{3}{c|}{\textbf{Multi-Document QA}} & \multicolumn{3}{c|}{\textbf{Summarization}} & \multicolumn{3}{c|}{\textbf{Few-shot Learning}} & \multicolumn{2}{c|}{\textbf{Synthetic}} & \multicolumn{2}{c|}{\textbf{Code}} &  \\
\cline{2-17}  
\makecell[c]{\textbf{KV}\\\textbf{Sparsity}}
& \makecell{\rotatebox{60}{\textbf{NtrvQA}}}
& \makecell{\rotatebox{60}{\textbf{Qasper}}}
& \makecell{\rotatebox{60}{\textbf{MF-en}}}
& \makecell{\rotatebox{60}{\textbf{HotpotQA}}}
& \makecell{\rotatebox{60}{\textbf{2WikiMQA}}}
& \makecell{\rotatebox{60}{\textbf{Musique}}}
& \makecell{\rotatebox{60}{\textbf{GovReport}}}
& \makecell{\rotatebox{60}{\textbf{QMSum}}}
& \makecell{\rotatebox{60}{\textbf{MultiNews}}}
& \makecell{\rotatebox{60}{\textbf{TREC}}}
& \makecell{\rotatebox{60}{\textbf{TrivialQA}}}
& \makecell{\rotatebox{60}{\textbf{SAMSum}}}
& \makecell{\rotatebox{60}{\textbf{PCount}}}
& \makecell{\rotatebox{60}{\textbf{PRe}}}
& \makecell{\rotatebox{60}{\textbf{Lec}}}
& \makecell{\rotatebox{60}{\textbf{RBP}}}
% & \textbf{}  \\
& \makecell[c]{\textbf{Avg.}} \\
\midrule
\multicolumn{18}{c}{\textbf{Llama-3-8B-Instruct}} \\
Dense         & 23.39 & 43.38 & 43.22 & 46.39 & 38.66 & 23.22 & 29.91 & 22.56 & 27.77 & 74.5  & 90.28 & 42.11 & 4.50  & 70.00 & 57.11 & 54.05 & \textbf{43.19} \\
K0.8 V0.0          & 22.67 & 39.08 & 39.44 & 44.98 & 38.51 & 21.94 & 21.75 & 21.00 & 23.88 & 69.00  & 90.24 & 36.92 & 7.50  & 64.50  & 49.15 & 45.79 & \textbf{39.77} \\
K0.9 V0.0         & 19.90 & 28.92 & 35.21 & 41.56 & 30.77 & 18.89 & 11.78 & 18.40 & 14.95 & 39.50  & 81.79 & 29.18 & 2.75 & 61.50  & 40.30 & 33.46 & \textbf{31.80} \\
K0.0 V0.8         & 24.48 & 42.54 & 43.96 & 45.48 & 38.71 & 22.46 & 24.47 & 21.64 & 25.09 & 73.00  & 90.11 & 39.03 & 5.62 & 64.00  & 56.39 & 56.54 & \textbf{42.22} \\
K0.0 V0.9          & 24.12 & 37.90 & 42.53 & 44.68 & \cellcolor{pink}38.29 & 21.99 & \cellcolor{yellow}20.22 & 21.29 & 21.61 & 69.00  & 90.15 & 36.04 & 3.29 & 62.50  & 55.87 & 53.59 & \textbf{40.19} \\
K0.8, V0.8    & 21.82 & 36.53 & 38.61 & 44.38 & 36.31 & 21.33 & 19.18 & 20.74 & 20.80  & 59.50  & 88.27 & 32.68 & 5.25 & 64.00  & 51.03 & 48.29 & \textbf{38.05} \\
K0.9, V0.9    & 17.47 & 24.13 & 30.64 & 38.63 & 29.24 & 17.24 & 13.50 & 19.67 & 15.03 & 35.50  & 75.29 & 27.39 & 5.50  & 63.00  & 41.77 & 34.39 & \textbf{30.52} \\
\bottomrule
\end{tabular}%
}
\end{table}

\section{Comparison with Semi-structured Sparsity}
\label{sec:appendix_2_4}
%\subsection{Structured-Pruned Key, Unstructured Value}
%\section{Joint Applications}
Between the structured pruning of rows and columns, and unstructured pruning of element, lies the 2:4 semi-structured sparsity where 2 out of 4 consecutive elements are non-zero, enforcing a global 50\% sparsity. Supported by NVIDIA Sparse Tensor Cores, 2:4 semi-structured sparsity also pursue the same objectives of Mustafar bitmap-based sparse format (Figure~\ref{fig:sub_bmp}), maximal compression and fast computation. In Table~\ref{tab:2_4}, we apply 2:4 semi-structured pruning to the per-token magnitude-based scheme. 
Comparing semi-structured sparsity to Key, Value, and both Key and Value cache to unstructured sparsity of Mustafar, we see that unstructured sparsity constantly outperforms semi-structured pattern of the same sparsity. This emphasizes the impact of fine-grained unstructured sparsity of element-wise pruning in model accuracy retention.  

\begin{table}[ht]
\centering
\caption{Comparison of 2:4 semi-structured and unstructured sparsity with Llama-3-8B-Instruct on LongBench}
\label{tab:2_4}
\vspace{0.5em}
\resizebox{\textwidth}{!}{%
\begin{tabular}{c|ccc|ccc|ccc|ccc|cc|cc|c}
\toprule
%\makecell{\textbf{KV} \\ \textbf{Sparsity}} & \multicolumn{3}{c|}{\textbf{Single-Document QA}} & \multicolumn{3}{c|}{\textbf{Multi-Document QA}} & \multicolumn{3}{c|}{\textbf{Summarization}} & \multicolumn{3}{c|}{\textbf{Few-shot Learning}} & \multicolumn{2}{c|}{\textbf{Synthetic}} & \multicolumn{2}{c|}{\textbf{Code}} & \textbf{Avg.} \\
%\midrule  
 & \multicolumn{3}{c|}{\textbf{Single-Document QA}} & \multicolumn{3}{c|}{\textbf{Multi-Document QA}} & \multicolumn{3}{c|}{\textbf{Summarization}} & \multicolumn{3}{c|}{\textbf{Few-shot Learning}} & \multicolumn{2}{c|}{\textbf{Synthetic}} & \multicolumn{2}{c|}{\textbf{Code}} &  \\
\cline{2-17}  
\makecell[c]{\textbf{KV}\\\textbf{Sparsity}}
& \makecell{\rotatebox{60}{\textbf{NtrvQA}}}
& \makecell{\rotatebox{60}{\textbf{Qasper}}}
& \makecell{\rotatebox{60}{\textbf{MF-en}}}
& \makecell{\rotatebox{60}{\textbf{HotpotQA}}}
& \makecell{\rotatebox{60}{\textbf{2WikiMQA}}}
& \makecell{\rotatebox{60}{\textbf{Musique}}}
& \makecell{\rotatebox{60}{\textbf{GovReport}}}
& \makecell{\rotatebox{60}{\textbf{QMSum}}}
& \makecell{\rotatebox{60}{\textbf{MultiNews}}}
& \makecell{\rotatebox{60}{\textbf{TREC}}}
& \makecell{\rotatebox{60}{\textbf{TrivialQA}}}
& \makecell{\rotatebox{60}{\textbf{SAMSum}}}
& \makecell{\rotatebox{60}{\textbf{PCount}}}
& \makecell{\rotatebox{60}{\textbf{PRe}}}
& \makecell{\rotatebox{60}{\textbf{Lec}}}
& \makecell{\rotatebox{60}{\textbf{RBP}}}
% & \textbf{}  \\
& \makecell[c]{\textbf{Avg.}} \\
\midrule
\multicolumn{18}{c}{\textbf{Llama-3-8B-Instruct}} \\
Dense         & 23.39 & 43.38 & 43.22 & 46.39 & 38.66 & 23.22 & 29.91 & 22.56 & 27.77 & 74.5  & 90.28 & 42.11 & 4.50  & 70.00 & 57.11 & 54.05 & \textbf{43.19} \\
\midrule
K0.5 (2:4)        &   21.79 & 39.77 & 42.34 & 45.15 & 38.81 & 21.72 & 24.34 & 22.21 & 25.44 & 69.50 & 90.87 & 39.10 & 7.00 & 62.50 & 54.33 & 50.29 & \textbf{40.95} \\
K0.5 (Unstructured)   & 23.40 & 43.68 & 43.63 & 46.00 & 38.60 & 22.72 & 29.39 & 22.33 & 27.64 & 74.50 & 90.66 & 41.09 & 5.00 & 68.50 & 55.89 & 52.39 & \textbf{42.84} \\
\midrule
V0.5 (2:4)      &    23.69 & 42.72 & 43.94 & 45.48 & 39.42 & 22.78 & 28.51 & 22.53 & 26.66 & 73.50 & 90.31 & 40.92 & 4.50 & 68.00 & 58.35 & 55.68 & \textbf{42.94} \\
V0.5 (Unstructured) & 23.80 & 43.14 & 43.32 & 46.28 & 39.42 & 22.97 & 29.18 & 22.70 & 27.13 & 74.50 & 90.50 & 41.74 & 5.00 & 67.50 & 57.23 & 54.30 &\textbf{43.04} \\
\midrule
K0.5(2:4) V0.5(2:4)   & 22.32 & 39.42 & 42.64 & 45.45 & 38.25 & 21.52 & 23.41 & 21.82 & 24.38 & 69 & 91.04 & 39.59 & 7.5 & 62.5 & 55.02 & 50.41 & \textbf{40.89} \\
K0.5 V0.5 (Unstructured) & 23.40 & 46.63 & 42.98 & 46.28 & 39.27 & 23.13 & 28.29 & 22.78 & 27.07 & 74.00 & 90.58 & 39.97 & 5.00 & 67.00 & 55.54 & 53.46 & \textbf{42.65} \\
\bottomrule
\end{tabular}%
}
\end{table}

\newpage
\section{Sparse Attention Kernel Details}
\label{sec:appendix_kernel}
As a supplement to Section~\ref{sec:3}, we offer more detail onto the Mustafar sparse attention kernel, which accelerates memory-bound batch SpMV. 
\subsection{Load-as-Compressed, Compute-as-Dense Pipeline}

\begin{wrapfigure}{r}{0.5\textwidth}  
    \centering
    \includegraphics[width=0.50\textwidth]{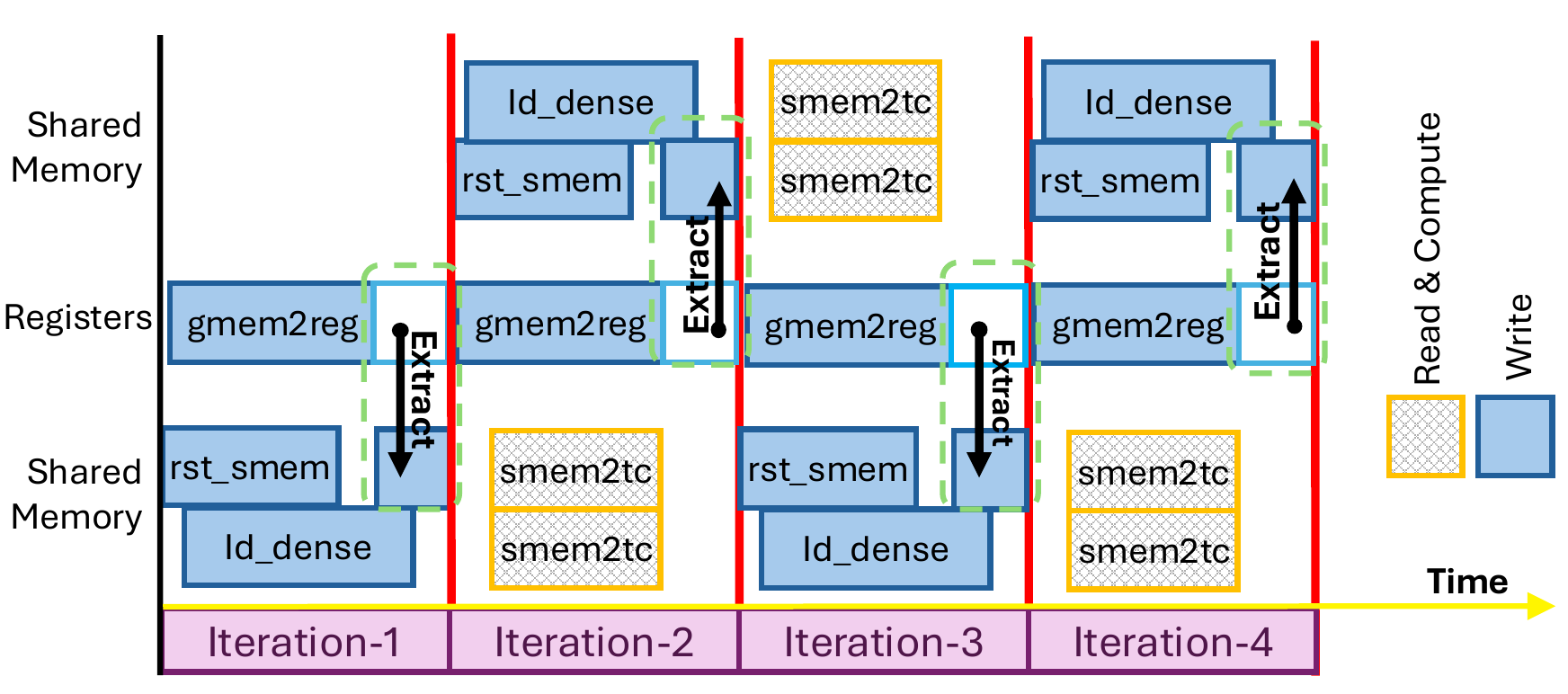}
    \caption{Load-as-compressed, compute-as-compute pipeline of FlashLLM~\cite{flashllm}}
    \label{fig:appendix_flashllm_pipeline}
\end{wrapfigure}

Crucial insight of accelerating SpMV involves reducing the data movement between the GPU global memory and the local memory of each GPU Streaming Multiprocessor. First proposed by FlashLLM~\cite{flashllm}, load-as-compressed, compute-as-dense pipeline as shown in Figure~\ref{fig:appendix_flashllm_pipeline} involves sending each matrix tile in the corresponding compressed form to the SM registers ('gmem2reg' in the figure), decompressing the compressed tile into the dense from to the shared memory ('extract'), then initializing computation on the next pipeline stage ('smem2tc'). Computation is mapped to tensor core to utilize the high fp16 compute throughput. To map MV, unused N dimensions are padded to zero for computation. Non-zero thread-tile of $1 \times 64$ in Figure~\ref{fig:sub_bmp} represents the granularity of non-zeros that a warp thread decompresses at a pipeline stage. Each warp thread decompresses 2 thread-tile per stage using the corresponding bitmap to determine the correct position of each non-zero. Effectively, each warp operates on a $64 \times 64$ sized matrix tile at a time. 

\subsection{KV Cache Management}
\begin{figure}[h]
\centering
\begin{subcaptionbox}{Tile Ordering of Key Cache\label{fig:appendix_key_tile}}[0.47\textwidth]
{\includegraphics[width=\linewidth]{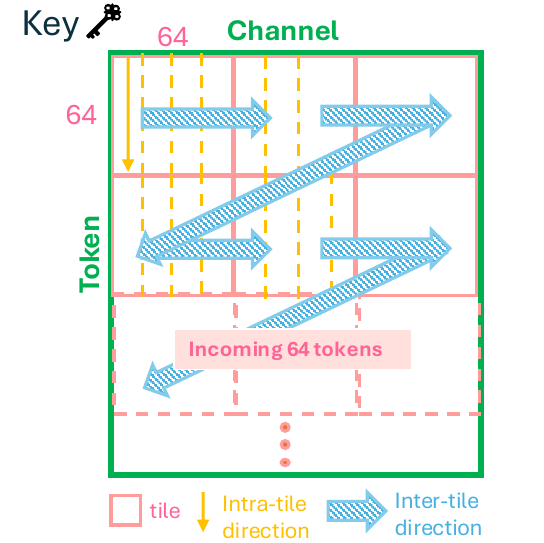}}
\end{subcaptionbox}
\hspace{0.03\textwidth}
\begin{subcaptionbox}{Tile Ordering of Value Cache\label{fig:appendix_value_tile}}[0.47\textwidth]
{\includegraphics[width=\linewidth]{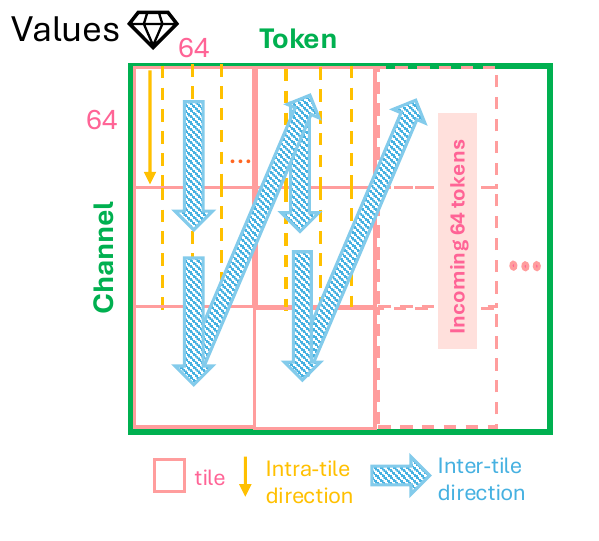}}
\end{subcaptionbox}
\caption{Tile ordering scheme of Key and Value cache} 
\label{fig:appendix_tiling}
\end{figure}

Tile size of $64 \times 64$ of each warp-tile (pink tiles in Figure~\ref{fig:appendix_tiling}), requires the KV cache to be compressed and appended to the existing KV cache in token groups of 64. Due to the dynamic nature of KV cache where new entries are added during generation, a kernel-compatible management of KV cache update is necessary. That is, (1) column tiling direction of KV cache must be orthogonal to the dimension that is being multiplied with: Key cache is multiplied on the channel-dimension, thus column tiling is across token dimension (yellow arrow in Figure~\ref{fig:appendix_key_tile}), value cache is multiplied on the token-dimension, thus column-tiling is across the channel dimension (yellow arrow in Figure~\ref{fig:appendix_value_tile}).  

(2), the layout of warp-tile must ensure that newly compressed tokens' KV cache can be appended to the existing compressed KV cache.
%Within a $64 \times 64$ sized warp-tile (pink tile in Figure~\ref{fig:appendix_tiling}), $1\times64$ sized thread-tiles are laid out in row-major order. 
As newly compressed KV cache are added onto the token dimension, traversal across multiple warp-tiles is done along channel-major dimension for both Key and and Value caches so that the compressed KV cache of the new tokens can be appended at the end. 

\subsection{Decode Speed Evaluation}
\label{sec:appendix_kernel_eval}

Extrapolating on Figure~\ref{fig:tput}, we evaluate Mustafar decoding on various input:output token ratios with batch size 4.
For Llama-2-7B, we use input sequence length of 2048. For Llama-3-8B-Instruct, we use input sequence length of 4096. We use output sequence lengths of 512, 1024, and 2048.

\begin{table}[h]
\centering
\caption{Decode speed comparison with dense inference}
\label{tab:decode_speed}
\vspace{0.5em}
\resizebox{\textwidth}{!}{%
\begin{tabular}{c|c|c|ccc}
\toprule
\textbf{Model} & 
\textbf{KV Format} & 
\textbf{TTFT} & 
\makecell{\textbf{Decode Speed}\\\textbf{(decode 512)}} &
\makecell{\textbf{Decode Speed}\\\textbf{(decode 1024)}} &
\makecell{\textbf{Decode Speed}\\\textbf{(decode 2048)}} \\
\midrule
Llama2 & Dense & 1.396 sec & 88.685 tokens / sec & 88.512 tokens / sec & 79.185 tokens / sec \\
       & Mustafar K0.5 V0.5 & 2.532 sec & 89.452 tokens / sec & 89.514 tokens / sec & 85.687 tokens / sec \\
       & Mustafar K0.7 V0.7 & 2.249 sec & 96.386 tokens / sec & 97.436 tokens / sec & 95.120 tokens / sec \\
\midrule
Llama3 & Dense & 2.769 sec & 61.993 tokens / sec & 61.220 tokens / sec & 59.242 tokens / sec \\
       & Mustafar K0.5 V0.5 & 3.269 sec & 78.434 tokens / sec & 83.768 tokens / sec & 83.303 tokens / sec \\
       & Mustafar K0.7 V0.7 & 3.151 sec & 84.065 tokens / sec & 88.293 tokens / sec & 89.699 tokens / sec \\
\bottomrule
\end{tabular}%
}
\end{table}

While Figure~\ref{fig:tput} measured the token throughput by considering both input and output tokens processed, in Table~\ref{tab:decode_speed} we derived the average decoding speed by measuring the end-to-end duration, and dividing it to the number of tokens generated to penalize Mustafar with the overhead of KV cache pruning and compression in both prefill and decode stages.

While time-to-first-token is delayed due to the overhead of pruning and compressing the KV cache during the prefill stage, the delay is offset by the accelerated attention computation during decoding, resulting in higher overall token generation throughput. Notably, Llama-3 exhibits a larger performance gain compared to Llama-2, as its GQA architecture reduces the overhead of KV cache pruning and compression.

\end{document}